\documentclass[acmsmall]{acmart}

\AtBeginDocument{%
  \providecommand\BibTeX{{%
    \normalfont B\kern-0.5em{\scshape i\kern-0.25em b}\kern-0.8em\TeX}}}

\setcopyright{acmcopyright}
\copyrightyear{2020}
\acmYear{2020}
\acmDOI{10.1145/1122445.1122456}

\acmJournal{CSUR}
\acmVolume{37}
\acmNumber{4}
\acmArticle{111}
\acmMonth{8}

\usepackage{subfig}
\usepackage{multirow}
\usepackage{makecell}
\usepackage{lineno}
\usepackage{ulem}

\begin{document}
% \linenumbers  %

%%
%% The "title" command has an optional parameter,
%% allowing the author to define a "short title" to be used in page headers.
\title{A Comprehensive Survey of Neural Architecture Search: Challenges and Solutions}

%%
%% The "author" command and its associated commands are used to define
%% the authors and their affiliations.
%% Of note is the shared affiliation of the first two authors, and the
%% "authornote" and "authornotemark" commands
%% used to denote shared contribution to the research.
\author{Pengzhen Ren}
\authornote{Both authors contributed equally to this research.}
\email{pzhren@foxmail.com}
\orcid{1234-5678-9012}
\author{Yun Xiao}
\authornotemark[1]
\email{yxiao@nwu.edu.cn}
\affiliation{%
  \institution{Northwest University}
}

\author{Xiaojun Chang}
\email{cxj273@gmail.com}
\affiliation{%
  \institution{Monash University}
}

\author{Po-Yao Huang}
\affiliation{%
	\institution{Carnegie Mellon University}
}

\author{Zhihui Li}
\authornote{Corresponding author.}
\affiliation{%
	\institution{Qilu University of Technology (Shandong Academy of Sciences)}
}

\author{Xiaojiang Chen}
\author{Xin Wang}
\affiliation{%
	\institution{Northwest University}
}

\renewcommand{\shortauthors}{Ren and Chang, et al.}

%%
%% The abstract is a short summary of the work to be presented in the
%% article.
\begin{abstract}
Deep learning has made breakthroughs and substantial in many fields due to its powerful automatic representation capabilities. It has been proven that neural architecture design is crucial to the feature representation of data and the final performance. 
However, the design of the neural architecture heavily relies on the researchers' prior knowledge and experience. And due to the limitations of human' inherent knowledge, it is difficult for people to jump out of their original thinking paradigm and design an optimal model. Therefore, an intuitive idea would be to reduce human intervention as much as possible and let the algorithm automatically design the neural architecture.
\textit{Neural Architecture Search} (NAS) is just such a revolutionary algorithm, and the related research work is complicated and rich. Therefore, a comprehensive and systematic survey on the NAS is essential. Previously related surveys have begun to classify existing work mainly based on the key components of NAS: search space, search strategy, and evaluation strategy. While this classification method is more intuitive, it is difficult for readers to grasp the challenges and the landmark work involved. Therefore, in this survey, we provide a new perspective: beginning with an overview of the characteristics of the earliest NAS algorithms, summarizing the problems in these early NAS algorithms, and then providing solutions for subsequent related research work. Besides, we conduct a detailed and comprehensive analysis, comparison, and summary of these works. Finally, we provide some possible future research directions.
\end{abstract}

%%
%% The code below is generated by the tool at http://dl.acm.org/ccs.cfm.
%% Please copy and paste the code instead of the example below.
%%


%%
%% Keywords. The author(s) should pick words that accurately describe
%% the work being presented. Separate the keywords with commas.

%%
%% This command processes the author and affiliation and title
%% information and builds the first part of the formatted document.
\maketitle

\section{Introduction}

Deep learning has already exhibited strong learning capabilities in many fields, including machine translation \cite{LSTM,The best of both worlds,Google's neural machine translation system}, image recognition \cite{VGG,Mobilenets,Imagenet classification with deep convolutional neural networks} and object detection \cite{Faster r-cnn,Ssd,Focal loss for dense object detection}. This is mainly due to the powerful automatic feature extraction capabilities offered by deep learning for unstructured data. Deep learning has transformed the traditional approach of manually designing features \cite{Histograms of oriented gradients for human detection,Object recognition from local scale-invariant features} into automatic extraction \cite{VGG,Resnet,GoogLeNet}, which has allowed researchers to focus on the design of neural architecture \cite{Neural architecture search with reinforcement learning,ENAS,MetaQNN}. However, designing neural architecture heavily relies on the researchers' prior knowledge and experience; this makes it difficult for beginners to make reasonable modifications to the neural architecture in line with their actual needs. Besides, people's existing prior knowledge and fixed thinking paradigms are likely to limit the discovery of new neural architectures to a certain extent. As a result, \textit{Neural architecture search} (NAS) was developed. 

NAS aims to design a neural architecture that achieves the best possible performance using limited computing resources in an automated way with minimal human intervention~\cite{ChengZHDCLDG20,ZhangLPCZGS2020}. NAS-RL \cite{Neural architecture search with reinforcement learning} and MetaQNN \cite{MetaQNN} are considered pioneers in the field of NAS. The neural architectures obtained by these works using \textit{reinforcement learning} (RL) methods have reached state-of-the-art classification accuracy on image classification tasks. This demonstrates that automated neural architecture design is feasible. Subsequently, the work of Large-scale Evolution \cite{Large-scale Evolution} once again verified the feasibility of this concept by using evolutionary learning to achieve similar results. However, these methods consume hundreds of GPU days or even more computing resources. This huge amount of computation is almost catastrophic for everyday researchers. Therefore, a large body of work has emerged on how to reduce the amount of calculation and accelerate the search for neural architecture \cite{Single path one-shot neural architecture search with uniform sampling,Understanding Architectures Learnt by Cell-based Neural Architecture Search,ENAS,Accelerating neural architecture search using performance prediction,Teacher guided architecture search,MdeNAS,AtomNAS,Simple and efficient architecture search for convolutional neural networks}. As the efficiency of search has improved, NAS has also been rapidly applied in the fields of object detection \cite{Detnas,Nas-fpn,Efficient Neural Architecture Transformation Search in Channel-Level for Object Detection,Computation Reallocation for Object Detection}, semantic segmentation \cite{Auto-deeplab,Fast neural architecture search of compact semantic segmentation models via auxiliary cells,Customizable architecture search for semantic segmentation}, adversarial learning \cite{Autogan}, speech recognition \cite{AutoSpeech}, architectural scaling \cite{Learnable embedding space for efficient neural architecture compression,Efficientnet,TAS}, multi-objective optimization \cite{ Efficient multi-objective neural architecture search via lamarckian evolution,Resource Constrained Neural Network Architecture Search,Nsga-net}, platform-aware \cite{Chamnet,ProxylessNAS,Mnasnet,Dpp-net}, data augmentation \cite{Autoaugment,Adversarial AutoAugment} and so on. Also, some works have considered how to strike a balance between performance and efficiency \cite{Partial order pruning,Sparse}. Although NAS-related research has been highly abundant, it is still difficult to compare and reproduce NAS methods \cite{NAS evaluation is frustratingly hard, NAS_Bench_1shot1, NAS_Bench, Nas-bench-101}.  This is because different NAS methods vary widely in terms of search space, hyperparameters, tricks, etc. Some works have also been devoted to providing a unified evaluation platform for popular NAS methods \cite{Nas-bench-101,NAS_Bench}.

\subsection{Motivation}
Due to the deepening and rapid development of NAS-related research, some methods that were previously accepted by researchers have proven to be imperfect by new research, leading to the development of an improved solution. For example, early incarnations of NAS trained each candidate neural architecture from scratch during the architecture search phase, leading to a surge in computation \cite{Neural architecture search with reinforcement learning,MetaQNN}. ENAS \cite{ENAS}  proposes to accelerate the architecture search process using a parameter sharing strategy. 
Due to the superiority of ENAS in terms of search efficiency, the weight sharing strategy was quickly recognized and adopted by a large number of researchers \cite{SMASH,Autogan,CAS}. However, soon afterward, new research finds that the widely accepted weight sharing strategy is likely to lead to candidate architectures being inaccurately ranked \cite{Evaluating the search phase of neural architecture search}; this make the algorithm difficult to the optimal neural architecture from a large number of candidate architectures, thereby further deteriorating the performance of the neural architecture that is eventually selected. Shortly afterward, DNA \cite{Blockwisely Supervised Neural Architecture Search with Knowledge Distillation}  modularized the large search space of NAS into blocks, enabling the candidate architecture to be fully trained to reduce the representation shift problem caused by the weight sharing. Besides, GDAS-NSAS \cite{Overcoming Multi-Model Forgetting in One-Shot NAS with Diversity Maximization} proposes a \textit{Novelty Search based Architecture Selection} (NSAS) loss function to solve the problem of multi-model forgetting (i.e. when weight sharing is used to sequentially train a new neural architecture, the performance of the previous neural architecture is reduced) caused by weight sharing during the super network training process. Similar research clues are very common in the rapidly developing field of NAS research.

More concisely, this survey has the following motivations:
\begin{itemize}
    \item Previous surveys often use the basic components of NAS to associate NAS-related work, which makes it difficult for readers to grasp the research ideas of NAS-related work.
    \item NAS-related fields are developing rapidly, and related work is complex and rich. There are obvious connections between different works, and existing surveys have not conducted a detailed and clear analysis of these links.
\end{itemize}
So a comprehensive and systematic survey based on challenges and solutions is highly beneficial to NAS research. 

\subsection{Our Contributions and Related Surveys}
The contributions of this survey are summarized as follows:
\begin{itemize}
    \item As far as we know, this is the first comprehensive and systematic review from the perspective of NAS challenges and corresponding solutions.
    \item We conduct a comprehensive analysis and comparison on the performance of existing NAS-related work and the optimization strategies they adopted.
    \item We analyze multiple possible development directions of NAS and point out two issues worthy of vigilance. The hyperparameter search of NAS is also discussed. They are very beneficial to the development of NAS.
\end{itemize}

\begin{table}[!tp]
    \centering
    \caption{Comparison of article frameworks of different NAS-related surveys.}
    \begin{tabular}{|c|c|c|}
    \hline
      Surveys &\makecell{Classification\\ standard} &  Main frame     \\ \hline
     Survey 1 \cite{Neural Architecture Search: A Survey}  & Basic components &  \multirow{2}{*}{\makecell[l]{Search space, Search strategy and Evaluation strategy}} \\\cline{1-2}
     Survey 2 \cite{A survey on neural architecture search} & Basic components &   \\\hline
     Our & \makecell{Challenges \\and solutions} & 
     \makecell[l]{
     (Challenge $\rightarrow$ Solution:)\\
    Global search space $\rightarrow$ Modular search space \\
    Discrete search strategy $\rightarrow$ Continuous search strategy  \\
    Search from scratch $\rightarrow$ Neural architecture recycling\\ 
    Fully trained $\rightarrow$ Incomplete training
     }
     \\ \hline
    \end{tabular}
    \label{tab:NAS frame comparison}
\end{table}

Previous related surveys classify existing work mainly based on the basic components of NAS: search space, search strategy, and evaluation strategy \cite{Neural Architecture Search: A Survey,A survey on neural architecture search}. Although this classification method is more intuitive, it is not conducive to assisting readers in capturing the research clues. Therefore, in this survey, we first summarize the characteristics and corresponding challenges of the early NAS methods. Based on these challenges, we then summarized and categorized existing research to present readers with a comprehensive and systematic overview of the extant challenges and solutions. A more specific NAS-related survey framework comparison is shown in Table
\ref{tab:NAS frame comparison}. 

\subsection{Article Organization}
We first made an insightful summary of the characteristics of early NAS in Section \ref{sec:Characteristics of early NAS}. Then, in response to the challenges faced by early NAS, we conducted a comprehensive and systematic analysis of the optimization strategy used by NAS in Section \ref{sec:Optimization Strategy} according to the following four parts: modular search space, continuous search strategy, neural architecture recycling, and incomplete training. In Section 
\ref{sec:Performance comparison}, we conducted a comprehensive performance comparison of NAS related work. In Section \ref{sec:Future Directions}, we discussed the future direction of NAS in detail. In Section \ref{sec:Review Threats}, we explained the corresponding review threats. Finally, in Section \ref{sec:Summary and Conclusions}, we made a summary and conclusion of this survey.

\section{Characteristics of early NAS}
\label{sec:Characteristics of early NAS}
\begin{figure*}[!tp]
	\centering
	\includegraphics[width=0.95\linewidth]{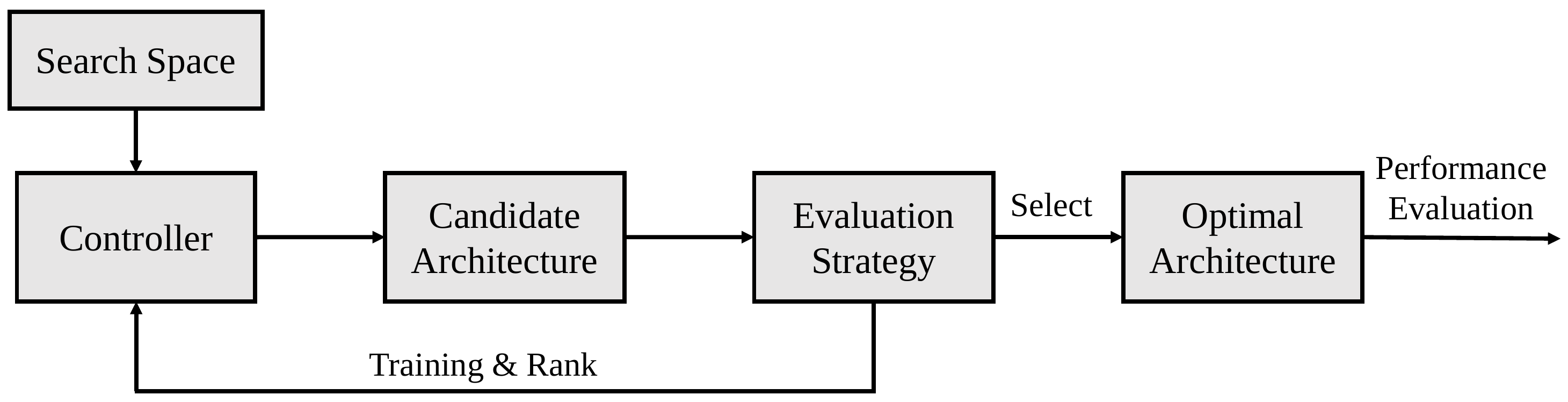}
	\caption{The general framework of NAS.}
	\label{fig:NAS_overall_framework}
\end{figure*}

In this section, we first analyze the early work of NAS and then summarize their overall framework and characteristics.The general framework of NAS is summarized in Fig.\ref{fig:NAS_overall_framework}. NAS generally begins with a set of predefined operation sets and uses a controller to obtain a large number of candidate neural architectures based on the search space created by these operation sets. The candidate architectures are then trained on the training set and ranked according to their accuracy on the validation set. The ranking information of the candidate neural architecture is then used as feedback information to adjust the search strategy, enabling a set of new candidate neural architectures to be obtained. When the termination condition is reached, the search process is terminated to select the best neural structure. The chosen neural architecture then conducts performance evaluation on the test set.

Early NAS also followed the above process to a large extent \cite{Neural architecture search with reinforcement learning,MetaQNN,Large-scale Evolution,GeNet}. The idea behind NAS-RL \cite{Neural architecture search with reinforcement learning} comes from the very simple observation that the architecture of a neural network can be described as a variable-length string. Therefore, an intuitive idea is that we can use RNN as a controller to generate such a string, then use RL to optimize the controller, and finally obtain a satisfactory neural architecture. MetaQNN \cite{MetaQNN} regards the selection process of the neural architecture as a Markov decision process, and uses \textit{Q}-learning to record rewards to obtain the optimal neural architecture. Large-scale Evolution \cite{Large-scale Evolution} aims to learn an optimal neural architecture automatically using \textit{evolutionary algorithms} (EA) while reducing human intervention as much as possible. This approach uses the simplest network structure to initialize a large population, then obtains the best neural architecture by reproducing, mutating, and selecting the population. GeNet \cite{GeNet}, which also uses EA, proposes a new neural architecture coding scheme, that represents the neural architecture as a fixed-length binary string. It randomly initializes a group of individuals, employs a predefined set of genetic operations to modify the binary string to generate new individuals, and finally selects the most competitive individual as the final neural architecture.

These early NAS approaches eventually made the automatically generated neural architecture a reality. To understand the reasons behind restricting the widespread use of early NAS, we have summarized the common characteristics existing in early NAS work from the perspective of a latecomer, as follows:
\begin{itemize}
\item \textbf{Global search space.} This requires the NAS to use a search strategy that searches all necessary components of the neural architecture. This means that NAS needs to find an optimal neural architecture within a very large search space. The larger the search space, the higher the corresponding search cost.

\item \textbf{Discrete search strategy.} This regards the differences between different neural architectures as a limited set of basic operations; that is, by discretely modifying an operation to change the neural architecture. This means that we cannot use the gradient strategy to quickly adjust the neural architecture. 

\item \textbf{Search from scratch.} In this approach, the model is built from scratch until the final neural architecture is generated. These methods ignore the existing neural architecture design experience and are unable to utilize the existing excellent neural architecture.

\item \textbf{Fully trained.} This approach requires training each candidate neural architecture from scratch until convergence. The network structures of the subsequent network and previous neural architectures are similar, as are those of the neural architectures at the same stage. Therefore, it is clear that this relationship not be fully utilized if each candidate neural architecture is trained from scratch. Also, we only need to obtain the relative performance ranking of the candidate architecture.  Whether it is necessary to train each candidate architecture to convergence is also a question worth considering.
\end{itemize}

To more clearly show the relationship between the characteristics of early NAS and NAS key components, we follow the expressions in the previous two NAS-related surveys \cite{A survey on neural architecture search,Neural Architecture Search: A Survey}, they generally regard search space, optimization methods, and performance estimation strategies as the three major components of NAS. In this paper, the global search space and full training of early NAS correspond to the two components of NAS search space and performance evaluation strategy respectively. The discrete search strategy and search from scratch in the early NAS correspond to the search strategy used in the RL and EA optimization methods in NAS.

\begin{figure}[!tp]
	\centering
	\includegraphics[width=0.4\linewidth]{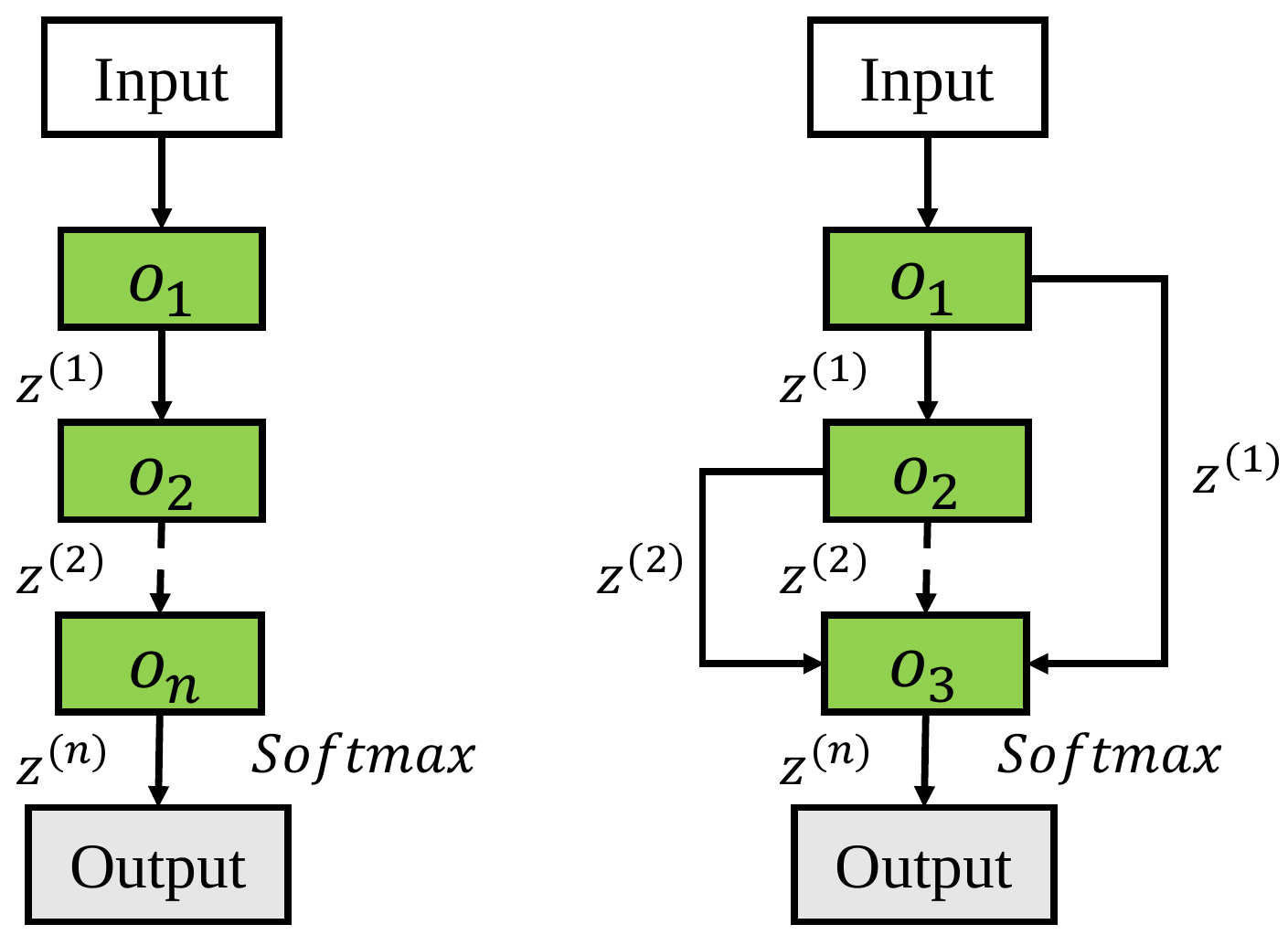}
	\caption{Two common global search spaces with chain structure in early NAS work. Left: The simplest example of a chain structure. Right: The chain structure example after adding skip connections. $o_i$ is an operation in the candidate set of operations and the $i$-th operation in the chain structure. The feature map generated by $o_i$ is represented as $z^{(i)}$. The input undergoes a series of operations to obtain the final output.}
	\label{fig:global_search_space}
\end{figure}

More specifically, the search space is determined by the predefined operation set and the hyperparameters of the neural architecture (for example architectural template, connection method, the number of convolutional layer channels used for feature extraction in the initial stage, etc.). These parameters define which neural architectures can be searched by the NAS. Fig.\ref{fig:global_search_space} presents examples of two common global search spaces with a chain structure in early NAS work. $o_i$ is an operation in the candidate operation set and the $i$-th operation in the chain structure. The feature map generated by $o_i$ is represented as $z^{(i)}$. The input undergoes a series of operations to obtain the final output. Fig.\ref{fig:global_search_space} (left): The simplest example of a chain structure MetaQNN \cite{MetaQNN}. At this point, for any feature map $z^{(i)}$, there is only one input node $z^{(i-1)}$, and $z^{(i)} = o_i\{(z^{(i-1)})\}$. Fig.\ref{fig:global_search_space} (right): The example after skip connections are added \cite{Neural architecture search with reinforcement learning,Large-scale Evolution,GeNet}. At this time, there can be multiple inputs for any feature map $z^{(i)}$, and
\begin{equation}
z^{(i)}=o^{(i)}\left(\left\{z^{(i-1)}\right\} 
\odot\left\{z^{(k)} | \alpha_{k, i}=1, k<i-1\right\}\right),
\end{equation}
where $\odot$ can be a sum operation or a merge operation; for example, $\odot$ is a merge operation in NAS-RL \cite{Neural architecture search with reinforcement learning}, and $\odot$ is an element-wise summation operation in GeNet \cite{GeNet}. It should be pointed out here that NASNet \cite{NASNet} considers these two operations in the experiment, but the experimental results demonstrate that the sum operation is better than the merge operation. Accordingly, since then, a large number of works have taken the summation operation as the connection method of the feature map obtained between different branch operations \cite{PNAS,AutoDispNet,DARTS}. Like the chain structure, Mnasnet \cite{Mnasnet} suggests searching for a neural architecture that is composed of multiple segments and connected in sequence, with each segment having its repeating structure.

\begin{figure}[!tp]
    \centering
    \includegraphics[width=0.4\linewidth]{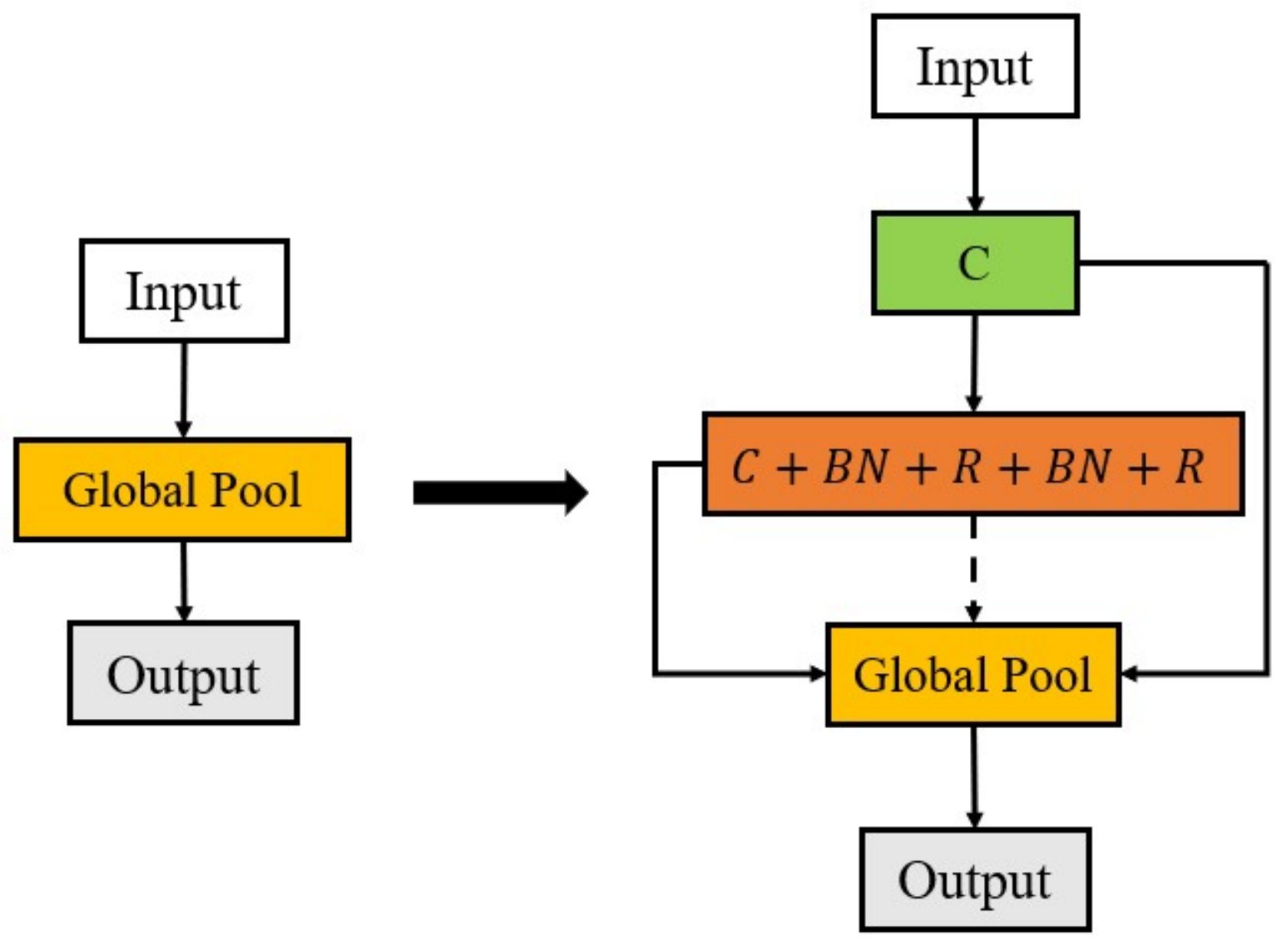}
    \caption{Taking Large-scale Evolution \cite{Large-scale Evolution} as an example, beginning from the simplest neural architecture to gradually generate the final neural architecture. $C$, $BN$ and $R$ denote the convolution, Batch Normalization and ReLU operations in sequence.}
    \label{fig:Search_from_scratch}
\end{figure}

Besides, in early NAS works, searching from scratch was a commonly adopted strategy. NAS-RL \cite{Neural architecture search with reinforcement learning} expresses the neural architecture as a string of variable length that is generated by RNN as a controller. The corresponding neural architecture is then generated according to the string, after which reinforcement learning is used as the corresponding search strategy to adjust the neural architecture search. MetaQNN \cite{MetaQNN} trains an agent to sequentially select the layer structure of the neural network on the search space constructed by the predefined operation set. This approach regards the layer selection process as a Markov decision process, and also uses \textit{Q}-learning as a search strategy to adjust the agent's selection behavior. Similar to NAS-RL \cite{Neural architecture search with reinforcement learning}, GeNet \cite{GeNet} also adopts the concept of encoding the network structure. The difference is that in GeNet \cite{GeNet}, the neural architecture representation is regarded as a string of fixed-length binary codes, which are regarded as the DNA of the neural architecture. The population is initialized randomly, after which evolutionary learning is used to reproduce, mutate and select the population, and then to iterate to select the best individual. It can be seen from the above analysis that these methods do not employ the existing excellent artificially designed neural architecture, but instead, search the neural architecture from scratch in their respective methods. More simply, Large-scale Evolution \cite{Large-scale Evolution} uses only a single-layer model without convolution as the starting point for individual evolution. Evolutionary learning methods are then used to evolve the population, and then to select the most competitive individuals in the population. We take Large-scale Evolution \cite{Large-scale Evolution} as an example and present an example of searching from scratch in Fig.\ref{fig:Search_from_scratch}.

The common characteristics of these early NAS works are also the collective challenges faced by the automatic generation of neural architecture. We summarize the solutions to the above-mentioned challenges in the subsequent NAS-related research work in Section \ref{sec:Optimization Strategy}.

\section{Optimization Strategy}
\label{sec:Optimization Strategy}

Regarding the characteristics and challenges of early NAS \cite{Neural architecture search with reinforcement learning,MetaQNN,Large-scale Evolution,GeNet}, in this section, we summarize the existing NAS research work regarding the following four aspects: 
\begin{itemize}
\item \textbf{\textit{Modular search space.}} Corresponding to the global search space, the modular search space treats the neural architecture as a stack of multiple different types of modules. Therefore, the search task is simplified from the original global search to only one or more modules of different types.
\item \textbf{\textit{Continuous search strategy.}} Corresponding to the discrete search strategy, the continuous search strategy continuously relaxes the structural parameters of the neural architecture so that they can be gradient optimized like network parameters.
\item \textbf{\textit{Neural architecture recycling.}} Corresponding to the search from scratch, neural architecture recycling takes the existing artificially designed high-performance neural architecture as a starting point and uses the NAS method to perform network transformations on them to improve their performance.
\item \textbf{\textit{Incomplete training.}} Corresponding to fully training, incomplete training aims to speed up the relative performance ranking of candidate architectures by making full use of the shared structure between candidate architectures or performance prediction, thereby avoiding resource consumption caused by complete training of all candidate architectures.
\end{itemize}

\subsection{Modular Search Space}
\label{sec:Modular search space}
Search space design has a critical impact on the final performance of the NAS algorithm. It not only determines the freedom of the neural architecture search but also directly determines the NAS algorithm's upper-performance limit to some extent. Therefore, the reconstruction of the search space is necessary.

One widely used approach is to transform the global search into a modular search space.  As a result, cell or block-based search space is commonly used in various NAS tasks because it can effectively reduce the complexity of NAS search tasks. This is mainly because the cell-based search space often needs to search only a few small cell structures, after which it repeatedly stacks such cells to form the final neural architecture. However, the global search space needs to search for all the components involved in building the entire neural architecture. Besides, the cell-based search space can be migrated to different data set tasks by stacking different numbers of cells, but this is often not possible when the global search space is used. Therefore, compared with the global search space, the cell-based search space is more compact and flexible.

\begin{figure}[!t]
	\centering
	\includegraphics[width=0.95\linewidth]{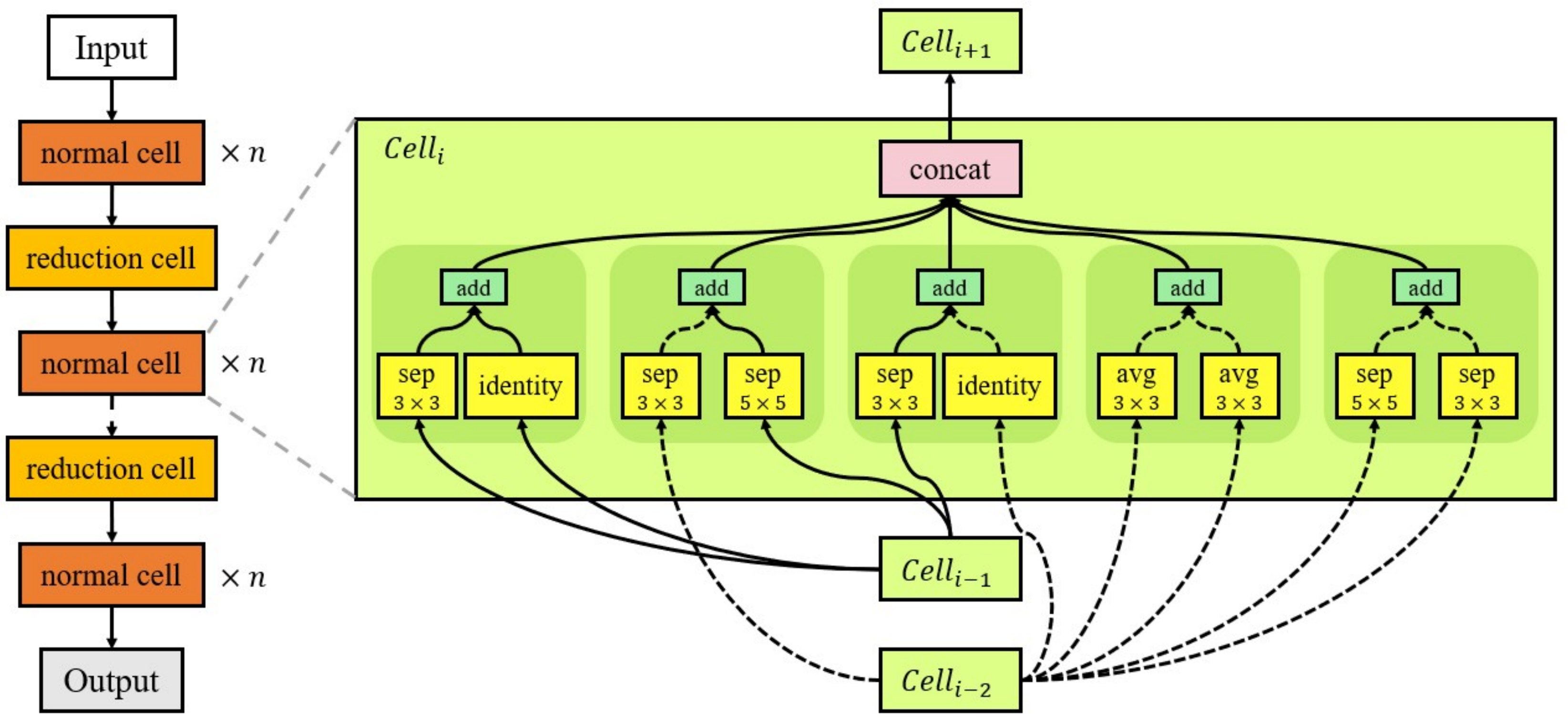}
	\caption{Left: The structure of the search space example based on two cells in \cite{NASNet}. The normal cell is repeated $n$ times and then connected to a reduction cell. The basic operation of the normal cell has a stride of 1, and the size of the feature map remains unchanged before and after output. The basic operation stride of the reduction cell is 2, and the size of the feature map is halved. Right: The best normal cell with 5 blocks searched in \cite{NASNet}.}
	\label{fig:cell_based}
\end{figure}

This concept mainly stems from the observation of the excellent neural architectures that have been artificially designed in recent years \cite{VGG,Resnet,GoogLeNet}. These artificial neural architectures typically accomplish the construction of the overall neural architecture by repeatedly stacking a certain unit operation or a small structure. In the NAS context, this small repeating structure is often called a cell. The construction of cell-based neural architecture is based on this idea. Neural architecture constructed in this way is not only superior in terms of performance but also easy to generalize. NASNet \cite{NASNet} is one of the first works to explore this idea. It proposes to search for two types of cells, namely normal cells and reduction cells. Normal cells are used to extract advanced features while keeping the spatial resolution unchanged, and reduction cells are mainly used to reduce the spatial resolution. Multiple repeated normal cells are followed by a reduction cell; this connection is then repeated multiple times to form the final neural architecture. In Fig.\ref{fig:cell_based} (left), we illustrate this kind of neural architecture based on two cells. In Fig.\ref{fig:cell_based} (right), we present the internal structure of an optimal normal cell in NASNet \cite{NASNet}. The structures of the corresponding reduction cell and normal cell are similar; the difference is that the basic operation step of the reduction cell is 2. A large number of subsequent works \cite{DARTS,AmoebaNet-A,P-DARTS} have used a search space similar to NASNet \cite{NASNet}.

\begin{figure}[!tp]
	\centering
	\includegraphics[width=0.95\linewidth]{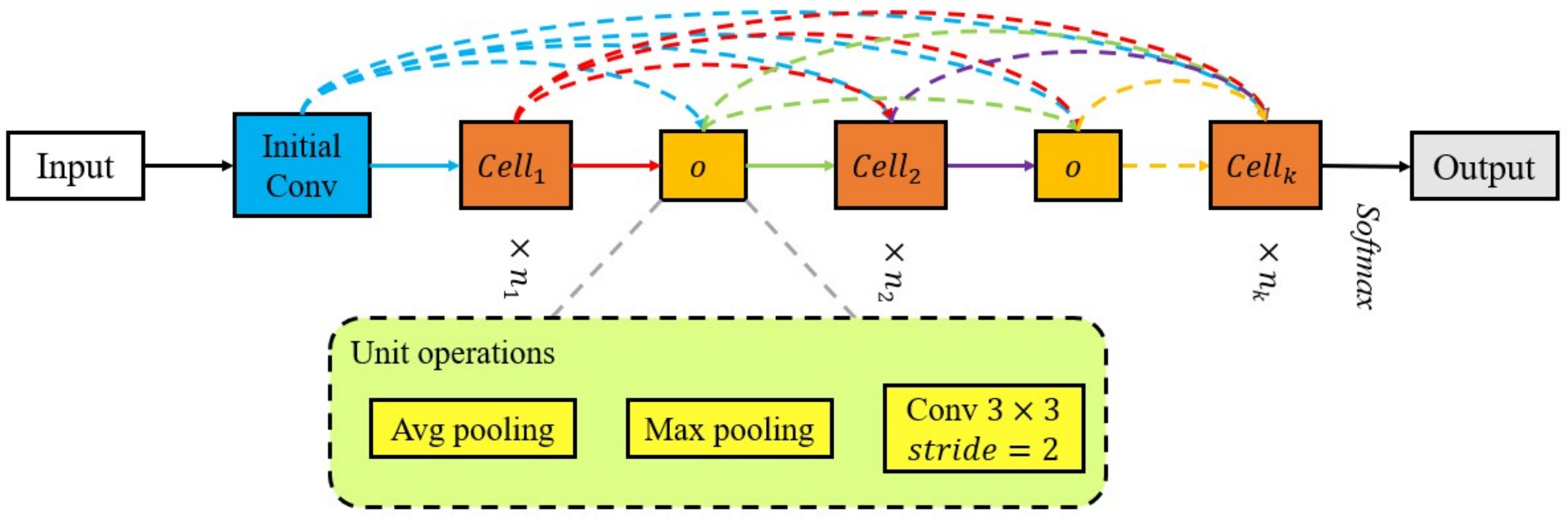}
	\caption{Using unit operations to replace the reduction cell in \cite{NASNet}, while having a densely connected neural architecture. The curved dotted line indicates the dense connection in Dpp-net \cite{Dpp-net}. The initial convolution is to extract low-level features. After the cell is repeated multiple times, a unit operation is used for downsampling. This connection is repeated multiple times to form the final neural architecture.}
	\label{fig:cell_based1}
\end{figure}

In ENAS \cite{ENAS}, its experiments provide strong evidence for the utilization of this similar cell-based search space. Subsequently, this cell-based search space is widely used in other research work. In \cite{Block-QNN,Hierarchical-EAS,Dpp-net,Evolving space-time neural architectures for videos}, to complete down-sampling, some unit operations are selected to replace reduction cell; at this time, the model only needs to search for a normal cell. We illustrate this structure in Fig.\ref{fig:cell_based1}. Here, the curved dotted line indicates the dense connection in Dpp-net \cite{Dpp-net}. At the same time as Block-QNN \cite{Block-QNN} of NASNet \cite{NASNet}, the pooling operation is used in place of the reduction cell to reduce the size of the feature map. Hierarchical-EAS \cite{Hierarchical-EAS} uses convolution with a kernel size of $3\times3$ and a stride of 2 instead of the reduction cell to reduce the spatial resolution. Furthermore, the idea of meta-operation is used to hierarchically build the cell structure. Dpp-net \cite{Dpp-net} is similar to Block-QNN \cite{Block-QNN}, but uses average pooling operation instead of a reduction cell. The difference is that Dpp-net \cite{Dpp-net} draws on the concept of DenseNet \cite{DenseNet} to use dense connections, including cells, to build a neural architecture, and further proposes to take devices into account for multi-objective optimization tasks. In \cite{Block-QNN,Hierarchical-EAS,Dpp-net}, the structure of each cell is the same, and it is only necessary to search for a cell. For video tasks, \cite{Evolving space-time neural architectures for videos} uses $L\times3\times 3$, $stride = 1,2,2$ max-pooling instead of a reduction cell. 

Moreover, to adapt to the complex task of video and expand the search space, the structure of each cell can be made different. AutoDispNet \cite{AutoDispNet} proposes to apply automatic architecture search technology in order to optimize large-scale U-Net-like encoder-decoder architectures. Therefore, it searches for three types of cells: normal, reduction, and upsampling. In the coding stage, the neural architecture comprises alternate connections of normal cells and reduction cells. In the decoding stage, it consists of a stack of multiple upsampling cells. \cite{Understanding Architectures Learnt by Cell-based Neural Architecture Search} studies the structural commonality of the cells obtained from some popular cell-based search spaces \cite{NASNet,AmoebaNet-A,ENAS,DARTS,SNAS} and defines the width and depth of cells. \cite{Understanding Architectures Learnt by Cell-based Neural Architecture Search} further proves theoretically and experimentally that due to the existence of the common connection mode, wide and shallow cells are easier to converge during training and easier to be searched, but the generalization effect is poor. This provides guidance that helps us to understand the cell-based NAS. Besides, there are many cell-based NAS-related works \cite{Autogan,SETN}.

In addition to repeatedly stacking one or more identical cells, FPNAS \cite{FPNAS} also considers the diversity of blocks when stacking blocks. The experimental results of FPNAS show that stacking diversified blocks is beneficial to the improvement of neural architecture performance, and FPNAS treats the search process as a bi-level optimization problem, which reduces the search cost to a level similar to that of the most advanced NAS method \cite{DARTS, SETN, ENAS}. Similar to FPNAS, FBNet \cite{Fbnet} explores a layer-wise search space. Specifically, FBNet fixes the macro-architecture and searches for blocks with multiple layers. Moreover, each block can have a different layer structure, and the blocks can also be different.

In this section, we conduct a comprehensive review of the modular search space. Compared with global search, the modular search space more effectively reduces the search space and makes NAS more accessible to researchers. Of course, this does not mean that the modular search space can meet all task requirements. Global search still has a unique research value because it provides the neural architecture design with a higher degree of freedom \cite{ProxylessNAS,Densely connected search space for more flexible neural architecture search}.

\subsection{Continuous Search Strategy}
\label{sec:Continuous Search Strategy}
NAS is regarded as a revolution in neural architecture design. However, NAS also requires high computational demand. For example, NASNet \cite{NASNet} uses the RL methods to spend 2000 GPU days to obtain the best architecture in CIFAR-10 and ImageNet. Similarly, AmoebaNet-A \cite{AmoebaNet-A} spends 3150 GPU days using evolutionary learning. One internal reason why these mainstream search methods based on RL \cite{Neural architecture search with reinforcement learning,MetaQNN,NASNet}, EA \cite{AmoebaNet-A,Large-scale Evolution}, Bayesian optimization \cite{NASBOT}, SMBO \cite{PNAS} and MCTS \cite{Deeparchitect} are so inefficient is that they regard neural architecture search as a black-box optimization problem in a discrete search strategy.

To address this issue, DAS \cite{Differentiable neural network architecture search} explores the possibility of transforming the discrete neural architecture space into a continuously differentiable form, and further uses gradient optimization techniques to search the neural architecture. This approach mainly focuses on the search of the hyperparameters of convolutional layers: filter sizes, number of channels, and grouped convolutions. MaskConnect \cite{Maskconnect} find that the existing cell-based neural architecture tends to adopt a predefined fixed connection method between modules; for example, each module only connects its first two modules \cite{Resnet}, or connects all previous modules \cite{DenseNet}. This connection method may not be optimal. Moreover, it uses the modified gradient method to explore the connection method between modules. Besides, other works \cite{Convolutional neural fabrics, Connectivity learning in multi-branch networks, Learning time/memory-efficient deep architectures with budgeted super networks} have also explored searching for neural architecture on continuous domains. However, the search for these neural architectures is limited to fine-tuning the specific structure of the network.

\begin{figure}[!tp] 
	\centering    %居中
	\subfloat[Connection to be determined] %第一张子图
    	{\centering          %子图居中
    		\includegraphics[width=0.2\linewidth]{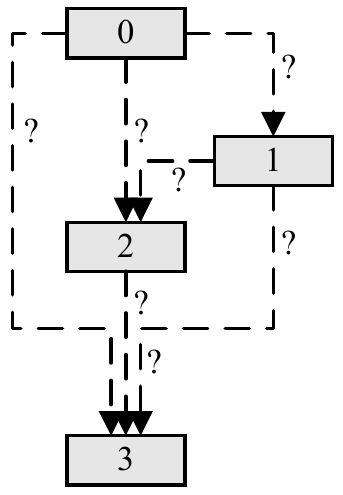}
    		\label{fig:DARTS_a}
    	}\hspace{3mm}
	\subfloat[Continuous relaxation] %第二张子图
    	{\centering      %子图居中
    		\includegraphics[width=0.2\linewidth]{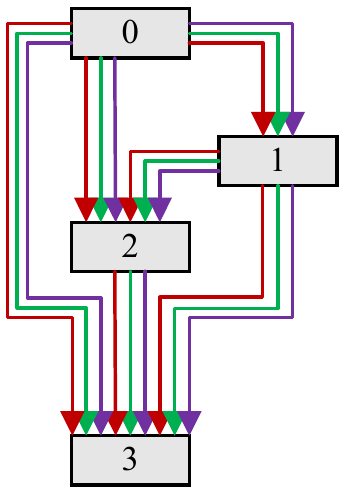}
    		\label{fig:DARTS_b}
    	}\hspace{3mm}
	\subfloat[Joint optimization] %第二张子图
    	{\centering          %子图居中
    		\includegraphics[width=0.2\linewidth]{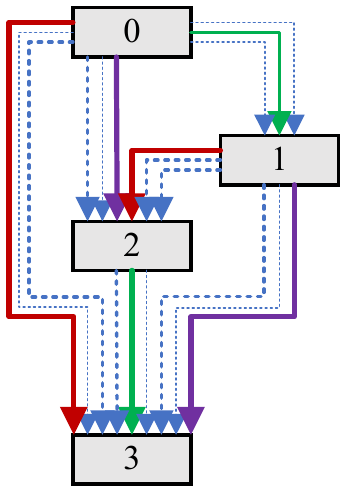}
    		\label{fig:DARTS_c}
    	}\hspace{3mm}
	\subfloat[Discrete neural architecture] %第二张子图
    	{\centering      %子图居中
    		\includegraphics[width=0.2\linewidth]{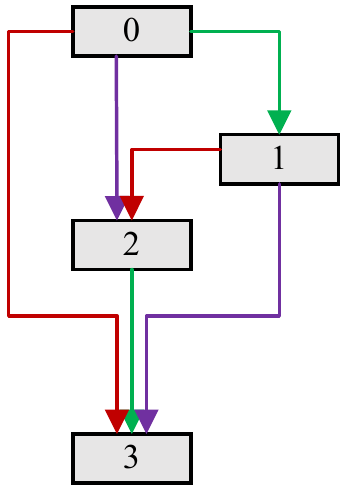}
    		\label{fig:DARTS_d}
    	}
	\caption{Continuous relaxation and discretization of search space in DARTS \cite{DARTS}. (a) A cell structure to be learned; the specific operation on the side is unknown. (b) Continuous relaxation of the cell-based search space, where each edge $e^{(i,j)}$ is a mixture of all candidate operations. (c) Joint optimization of the probability of mixed operations and network weights. (d) Discrete searched neural architecture.} 
	\label{fig:DARTS}  %图片引用标记
\end{figure}

In order to solve the above challenges, DARTS \cite{DARTS} was developed. DARTS continuously relaxes the originally discrete search strategy, which makes it possible to use gradients to efficiently optimize the architecture search space. DARTS follows the cell-based search space of NASNet \cite{NASNet} and further normalizes it. Every cell is regarded as a \textit{directed acyclic graph} (DAG), which is formed by sequentially connecting $ N $ nodes. Each of these cells has two input nodes and one output node. For convolutional cells, the input node is the output of the first two cells; for the recurrent cell, one is the input of the current step, while the other is the state of the previous step. The cell output is the concatenation result of all intermediate nodes. Each intermediate node $x^{(j)}$ in the cell is a potential feature representation, and is linked with each previous intermediate node $x^{(i)}$ in the cell through a directed edge operation $o^{(i,j)}$. For a discrete search strategy, each intermediate node can be expressed as follows:
\begin{equation}
	x^{(j)}=\sum_{i<j} o^{(i, j)}\left(x^{(i)}\right).
\end{equation}
The DARTS approach makes the discrete search strategy continuous by relaxing the selection of candidate operations to a \textit{softmax} of all possible operations. The mixed operation $\bar{o}^{(i, j)}(x)$ applied to feature map $x$ can be expressed as follows:
\begin{equation}\bar{o}^{(i, j)}(x)=\sum_{o \in \mathcal{O}} \frac{\exp \left(\alpha_{o}^{(i, j)}\right)}{\sum_{o^{\prime} \in \mathcal{O}} \exp \left(\alpha_{o^{\prime}}^{(i, j)}\right)} o(x)
\end{equation}
where $\mathcal{O}$ represents a set of candidate operations,  while $\alpha_{o}^{(i, j)}$ represents the weight of operation $o$ on directed edge $e^{(i,j)}$. Therefore, the neural architecture search has evolved into an optimization process for a set of continuous variables $\alpha=\{\alpha^{(i,j)}\}$. Once the search is complete, the most likely operation $o^{(i,j)}$ on the directed edge $e^{(i,j)}$ 
is selected while other operations is discarded.
\begin{equation}
	o^{(i, j)}=\operatorname{argmax}_{o \in \mathcal{O}} \alpha_{o}^{(i, j)}
\end{equation}
By solving a  bilevel optimization problem \cite{Hierarchical optimization,An overview of bilevel optimization}, the probability of mixed operations (the parameters $\alpha$ of the neural architecture) and network weights $w$ can be jointly optimized as follows:
\begin{equation}
	\begin{array}{cl}
		\min\limits_{\alpha} & \mathcal{L}_{v a l}\left(w^{*}(\alpha), \alpha\right) \\
		\text { s.t. } & w^{*}(\alpha)=\operatorname{argmin}_{w} \mathcal{L}_{train}(w, \alpha)
	\end{array}
\end{equation}
where $  \mathcal{L}_{v a l}$ and $\mathcal{L}_{train}$ denote validation and training losses, respectively, while $\alpha$ is the  upper-level variable and $w$ is the  lower-level variable. By jointly optimizing this problem, the optimal $\alpha$ is obtained, after which the discretization is performed to obtain the final neural architecture. We illustrate this process in Fig.\ref{fig:DARTS}.

Compared with DARTS, the neural architecture search process is changed from the selection of discrete candidate operations to the optimization of the probability of continuous mixed operations. During the same period, NAO \cite{NAO} opts to encode the entire neural architecture to map the originally discrete neural architecture to continuously embedded encoding. Subsequently, the output of the performance predictor is maximized by the gradient optimization method to enable the optimal embedded coding to be obtained. Finally, a decoder is used to discretize the optimal continuous representation (that is, the optimal embedded coding) into the optimal neural architecture. Furthermore, DARTS uses the $argmax$ strategy to eliminate the less probable operations among the mixed operations as a means of discretizing the neural architecture. However, common non-linear problems in network operation can introduce bias into the loss function; this bias exacerbates the performance difference between the derived child networks and the converged parent networks, which results in the need to retrain the parameters of the derived child networks. Therefore, a NAS solution with reduced performance deviation between the derived child networks and the converged parent networks is necessary. To this end, SNAS \cite{SNAS} begins with the delayed reward of reinforcement learning, then determines why delayed reward leads to the slow convergence speed of reinforcement learning when performing architecture search. Accordingly, SNAS proposes remodeling the NAS to theoretically bypass the delayed reward problem for reinforcement learning, while simultaneously ensuring that neural architecture parameters are continuous, so those network operation parameters and neural architecture parameters can be jointly optimized using a gradient method. Based on this, SNAS has a more efficient and automated neural architecture search framework that still maintains the completeness and differentiability of the NAS pipeline.

In works about SNAS, DARTS and many other NAS \cite{Nas-fpn,Convnet architecture search for spatiotemporal feature learning,Searching for efficient multi-scale architectures for dense image prediction,Nas-bench-101}, the feasible paths of the searched neural architecture depend on each other and are closely coupled during the search phase. While SNAS does to some extent reduce the performance difference between the derived child network and the converged parent network, SNAS and DARTS are still required to choose only one path during the verification phase. This crude decoupling inevitably leads to a gap between neural architectures during the search and verification phases. To address this issue, DATA \cite{DATA} developed the \textit{Ensemble Gumbel-Softmax} (EGS) estimator, which can decouple the relationship between different paths of the neural architecture and achieve the seamless transfer of gradients between different paths. This solves the problem of the architecture's inability to be seamlessly connected between search and verification. 

\begin{figure}[!tp] 
	\centering    %居中
	\subfloat[DARTS cell] %第一张子图
	{\centering          %子图居中
		\includegraphics[width=0.2\linewidth]{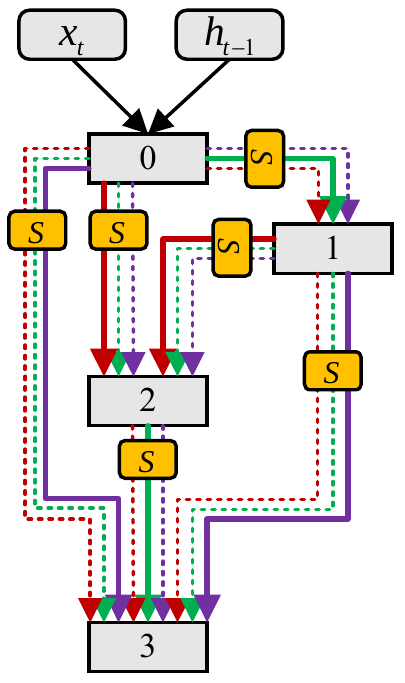}
		\label{fig:I-DARTS_a}
	}\hspace{10mm}
	\subfloat[I-DARTS cell] %第二张子图
	{\centering      %子图居中
		\includegraphics[width=0.2\linewidth]{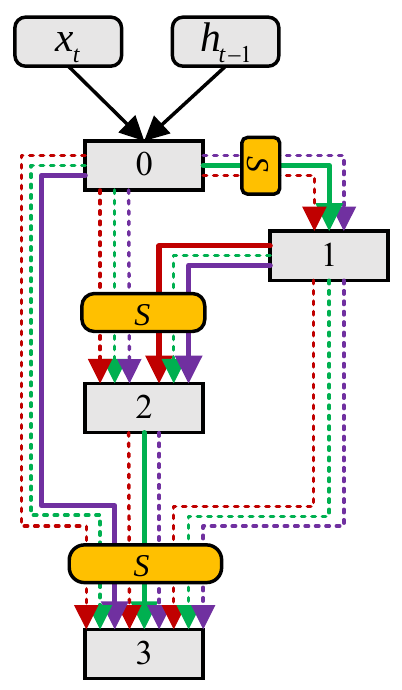}
		\label{fig:I-DARTS_b}
	}\hspace{3mm}
	{\centering      %子图居中
		\includegraphics[scale=0.7]{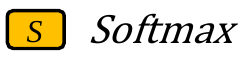}
	}
	\caption{Comparison of the cell structure of DARTS \cite{DARTS} and I-DARTS \cite{I-DARTS} in a recurrent neural network. (a) In DARTS, edges from different nodes cannot be compared. Moreover, when discretizing the neural architecture, there is only one correlation edge between each pair of nodes. (b) In I-DARTS, a given node uses a \textit{softmax} while considering all input edges. Given a specific node, I-DARTS can determine whether edges are connecting related nodes according to the importance of all input edges: either there are multiple connected edges or no related edges.} 
	\label{fig:I-DARTS}  %图片引用标记
\end{figure}

Furthermore, I-DARTS \cite{I-DARTS} notes that the \textit{softmax}-based relaxation constraint between each pair of nodes may cause DARTS to be a "local" model. In DARTS, the cell's middle node is connected to all the precursor nodes, and when discretizing the neural architecture, there is an only one-directional edge between each pair of nodes. This results in the existence of edges from different nodes that are unable to be compared with each other. Besides, DARTS requires only one directed edge between each pair of nodes; this constraint design has no theoretical basis and also limits the size of the DARTS search space. These local decisions, which are caused by the bias problem in graph-based models \cite{Conditional random fields,Probabilistic graphical models} compel DARTS unable to make the best choice of architecture. Based on this, I-DARTS proposes an interesting and simple idea: namely, using a \textit{softmax} to simultaneously consider all input edges of a given node. We present the cell structure comparison of DARTS and I-DARTS in the recurrent neural network in Fig.\ref{fig:I-DARTS}. As can be seen from Fig.\ref{fig:I-DARTS_b}, when given a node, I-DARTS can determine whether edges are connecting related nodes depending on the importance of all input edges: there are either multiple connected edges or no related edges. 

P-DARTS \cite{P-DARTS} starts with the deep gap between the search and evaluation of neural architecture, and improves DARTS overall. In DARTS, due to limitations on computational resources, DARTS uses a shallow cell stack architecture in the search phase; in the evaluation phase, moreover, it stacks more searched cells to enable the processing of data sets with higher resolution. Therefore, the basic cell of the neural architecture used for evaluation is designed for shallow architecture, which differs from the deep neural architecture used in the evaluation stage. Based on this, P-DARTS uses progressive search to gradually increase the depth of the network during the search phase. Moreover, it also gradually reduces the candidate operation set according to the weights of the mixed operation in the search process, to cope with the problem of the increase in the calculation volume caused by the increase in depth. At the same time, P-DARTS proposes regularization of the search space to deal with the problem of insufficient stability during the process of searching in deep architectures (algorithms are heavily biased towards skip-connect).

\begin{figure}[!tp]
	\centering
	\includegraphics[width=0.6\linewidth]{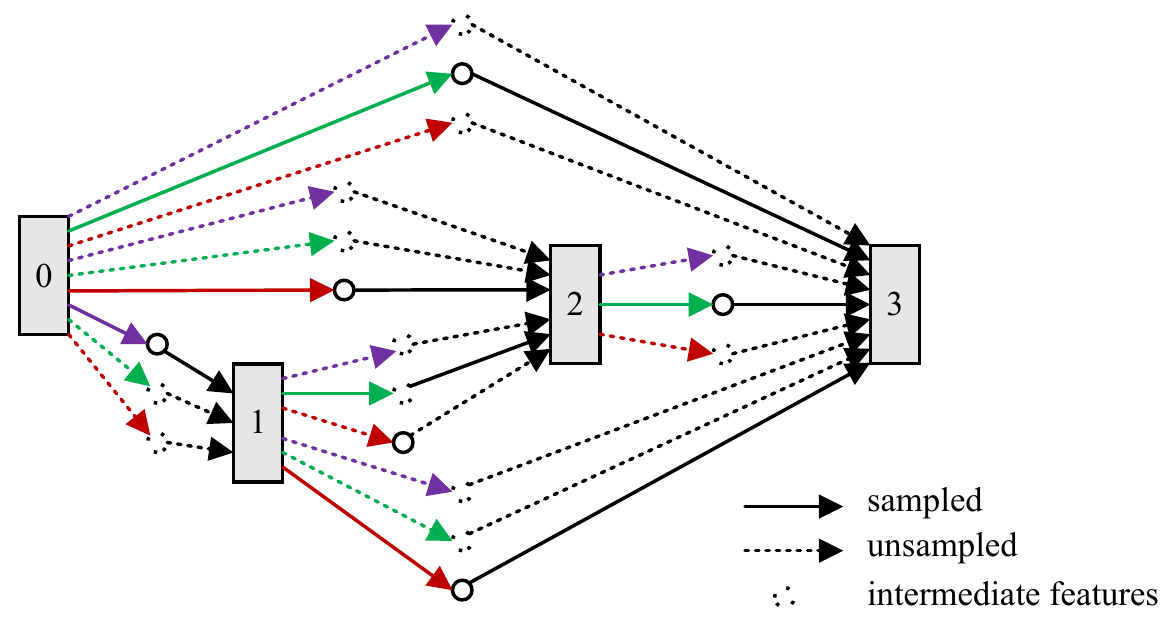}
	\caption{In GDAS \cite{GDAS}, an example of sampling subgraphs in a DAG with three intermediate nodes. Colored lines indicate operations in the candidate operation set, while black lines indicate corresponding information flows. The circles indicate the intermediate features following operation processing. Each intermediate node is equal to the sum of all the sampled intermediate features. In one iteration, only one sampled subgraph is trained.}
	\label{fig:GDAS}
\end{figure}

Compared with NAS based on RL and EA, DARTS greatly improves the search efficiency, but to an insufficient extent. As shown in Fig.\ref{fig:DARTS_c}, in the search phase, DARTS continuously relaxes the cell's search space and optimizes all parameters in the DAG simultaneously. This causes DARTS to occupy too much memory on the device during searching, which slows down the search speed. At the same time, the effects of different operations between the same pair of nodes may be canceled by each other, thereby destroying the entire optimization process. To this end, GDAS \cite{GDAS} proposes to use a differentiable architecture sampler in each training iteration to sample only one subgraph in the DAG, meaning that only one part of the DAG needs to be optimized in any one iteration. We illustrate this process in Fig.\ref{fig:GDAS}. At the same time, in the verification stage, the architecture sampler can be optimized by using the gradient-based method, thereby further improve the search efficiency of GDAS.

To reduce DARTS's memory usage during search and improve the search efficiency, PC-DARTS \cite{PC-DARTS} opts to start from the channel, as opposed to GDAS' sampling subgraphs in DAG and training only one subgraph in one iteration. During the search process, PC-DARTS samples the channels and convolves only the sampled channel features to achieve efficient gradient optimization. To deal with the problem of inconsistent information brought about by the channel sampling strategy, PC-DARTS uses edge normalization to solve this problem. It reduces the uncertainty in the search process by adding a set of edge-level parameters. As a result, PC-DARTS can save memory and is more efficient and stable. \cite{R-DARTS} recently find that DARTS \cite{DARTS} exhibits poor test performance for architecture generated in a wide search space. This work contends that when the discovered solutions are consistent with the high verification loss curvature in the architecture space, the discovered architecture is difficult to promote.  Moreover, various types of regularization are added to explore how to make DARTS more robust. Finally, \cite{R-DARTS} proposes several simple variants and achieve good generalization performance. Although we have conducted many reviews, there are still many improvements that have been made based on DARTS \cite{XNAS,SGAS}.

In the above-mentioned gradient-based methods, local optimization is a common problem. Therefore, we conduct a comprehensive review of the solution to this problem here. The experimental results of DARTS \cite{DARTS} show that an appropriate learning rate helps the model converge to a better local optimal value. As shown in Fig.\ref{fig:I-DARTS_b}, I-DARTS \cite{I-DARTS} relaxes the softmax-based relaxation on each edge of DARTS to all incoming edges of a given node, thereby alleviating the impact of bias caused by local decision-making. PC-DARTS \cite{PC-DARTS} uses channel sampling to replace the convolution operation on all channels in DARTS, thereby reduces the possibility of falling into a local optimum. In general, local optimization is still an important challenge faced by gradient-based optimization methods, so more related research is needed in the future.

In this section, we provide a comprehensive and systematic overview of optimized NAS work that employs gradient strategies on a continuous search strategy. Due to the simplicity and elegance of the DARTS architecture, the research work related to DARTS is quite rich. Moreover, gradient optimization in a continuous search strategy is an important trend of NAS.

\subsection{Neural Architecture Recycling}
\label{sec:Neural Architecture Recycle}
Early NAS works \cite{Neural architecture search with reinforcement learning,MetaQNN,Large-scale Evolution,GeNet} and many subsequent works \cite{DARTS, FPNAS, Resource Constrained Neural Network Architecture Search,Assemblenet} aim to search the neural architecture from scratch. From a certain perspective, this type of approach does increase the freedom of neural architecture design, and it is very likely to result in the design of a new high-performance network structure unknown to humans. However, it is clear that this idea also increases the time complexity of searching for the best neural architecture; this is because it does not make full use of the prior knowledge regarding the existing artificially designed high-performance neural architecture. Therefore, a new idea would be to use the existing, artificially designed high-performance neural architecture as a starting point, then use the NAS method to modify or evolve these neural architectures, as this would enable a more promising neural architecture to be obtained at a lower computing cost. This process is generally referred to as 'network transformation'. 

\begin{figure}[!tp]
	\centering
	\includegraphics[width=0.95\linewidth]{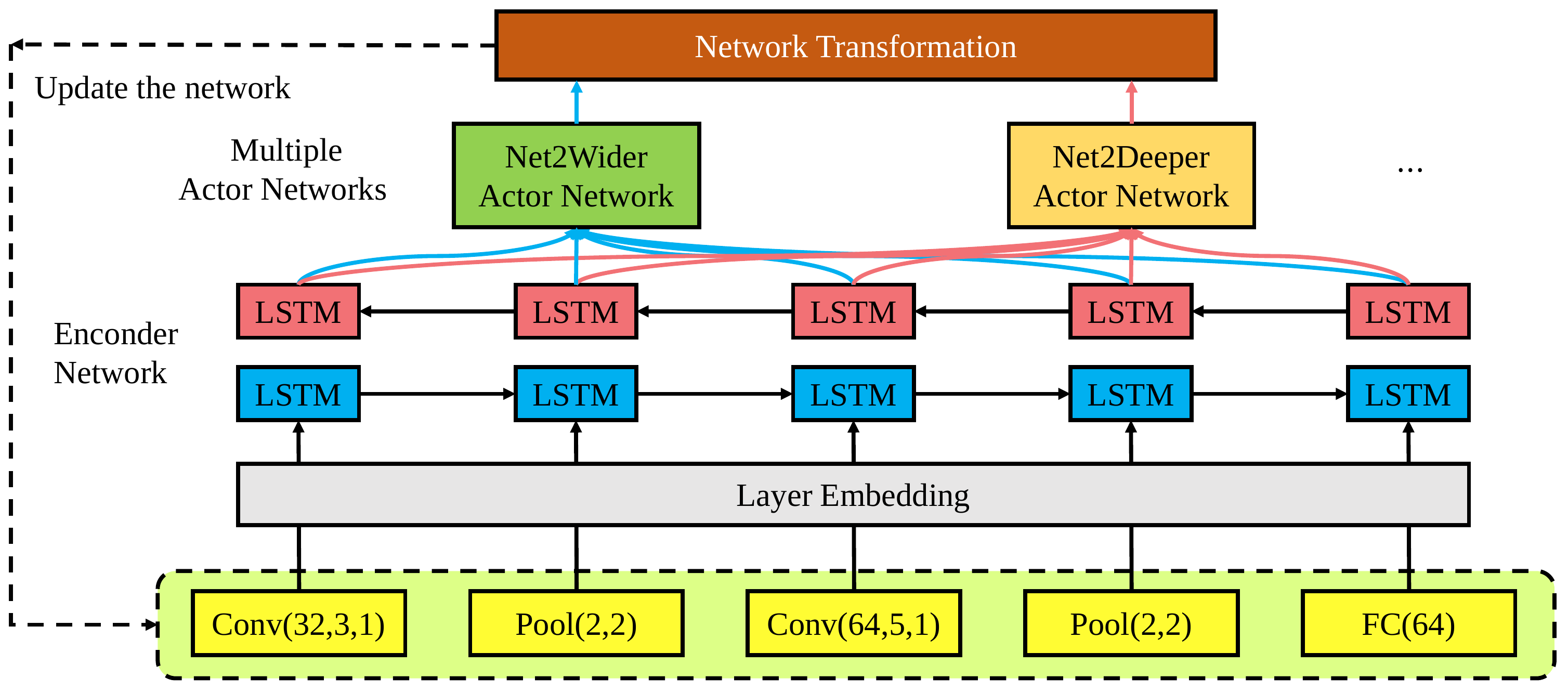}
	\caption{In EAS \cite{EAS}, architecture search based on network transformation. After the existing network layer is encoded by the layer encoder, Bi-LSTM \cite{Bi-LSTM} is used as a meta-controller to learn the low-dimensional feature representation of the neural architecture. The multi-actor network combines these features to decide whether the corresponding network transformation operation (deepening layer or widening layer) should be adopted. Finally, reinforcement learning is used to update the meta-controlled parameters.}
	\label{fig:EAS}
\end{figure}

\begin{figure*}[!tp]
    \centering
    \includegraphics[width=0.95\linewidth]{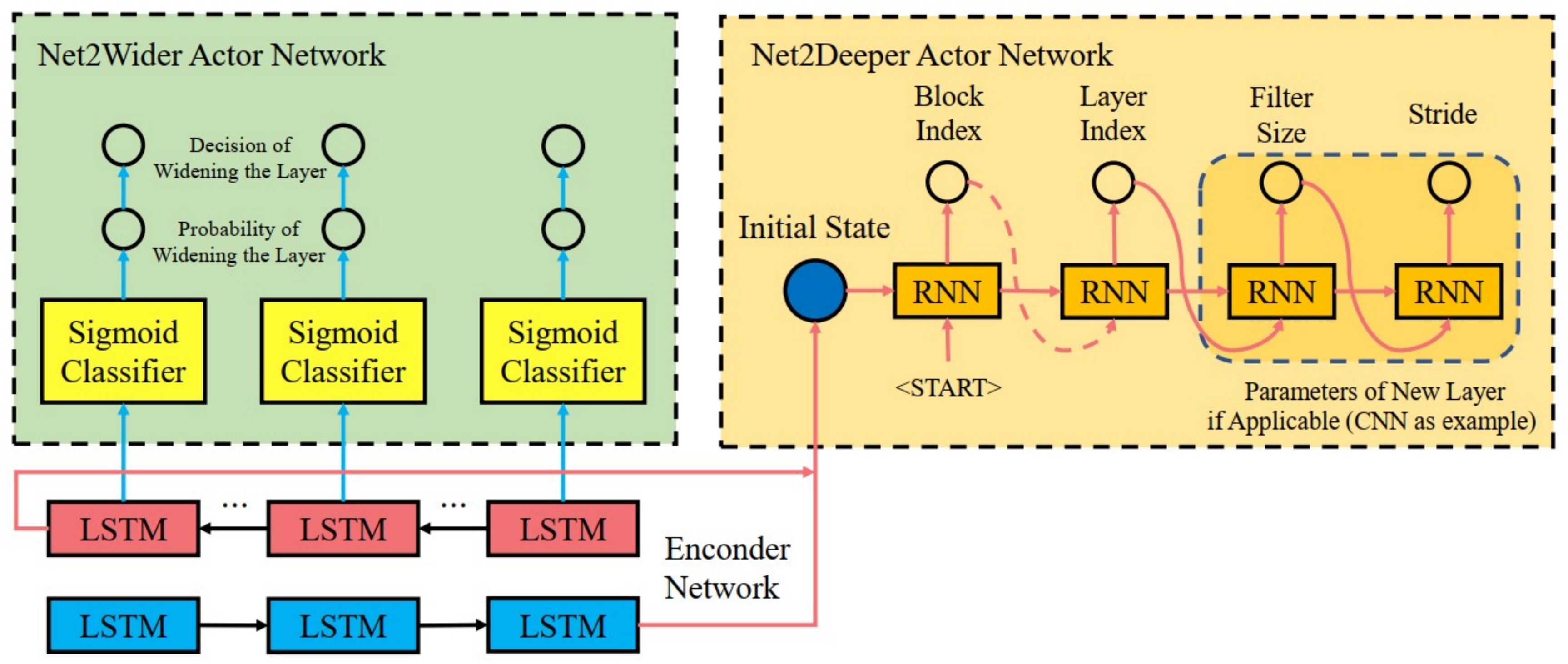}
    \caption{The internal structure of two actor networks in EAS \cite{EAS}: Net2Wider and  Net2Deeper. In Net2Wider, the actor network shares the same sigmoid classifier and decides whether to widen the layer according to each hidden state of the encoder. In Net2Deeper, the actor network inputs the state of the final hidden layer of Bi-LSTM into the recurrent network, after which the recurrent network decides where to insert the layer and the parameters of the inserted layer.}
    \label{fig:ActionNetwork}
\end{figure*}

Net2Net \cite{Net2Net} conducts a detailed study of network transformation technology and proposes function-preserving transformations to facilitate the reuse of model parameters after transformation. This approach can effectively accelerate the training of new and larger networks. Based on this idea, \cite{EAS} proposes \textit{efficient architecture search} (EAS), which uses the encoder network as a meta-controller to learn the low-dimensional representation of the existing neural architecture, and further refers to the multiple actor networks in Net2Net \cite{Net2Net} to decide whether to make corresponding adjustments to the neural architecture at the layer level (deepening or widening layer). Besides, this approach uses reinforcement learning strategies to update the parameters in the meta-controller. EAS takes the view that the network transformation at the layer level needs to combine the information of the entire neural architecture; thus, a \textit{bidirectional recurrent network} (Bi-LSTM) \cite{Bi-LSTM} is used as the network encoder. Since EAS is a network transformation on an existing network, models and weights can be reused to substantially reduce the amount of calculation required. We illustrate the overall neural architecture of EAS in Fig.\ref{fig:EAS}. In Fig.\ref{fig:ActionNetwork}, we also present the internal structure of two actor networks: Net2Wider and  Net2Deeper. In Net2Wider, the actor network shares the same \textit{sigmoid} classifier and decides whether to widen the layer according to each hidden encoder state. In Net2Deeper, the actor network inputs the state of the final hidden Bi-LSTM layer into the recurrent network, after which the recurrent network decides both where to insert the layer and the parameters of the inserted layer.

\begin{figure*}[!tp]
	\centering
	\includegraphics[width=0.95\linewidth]{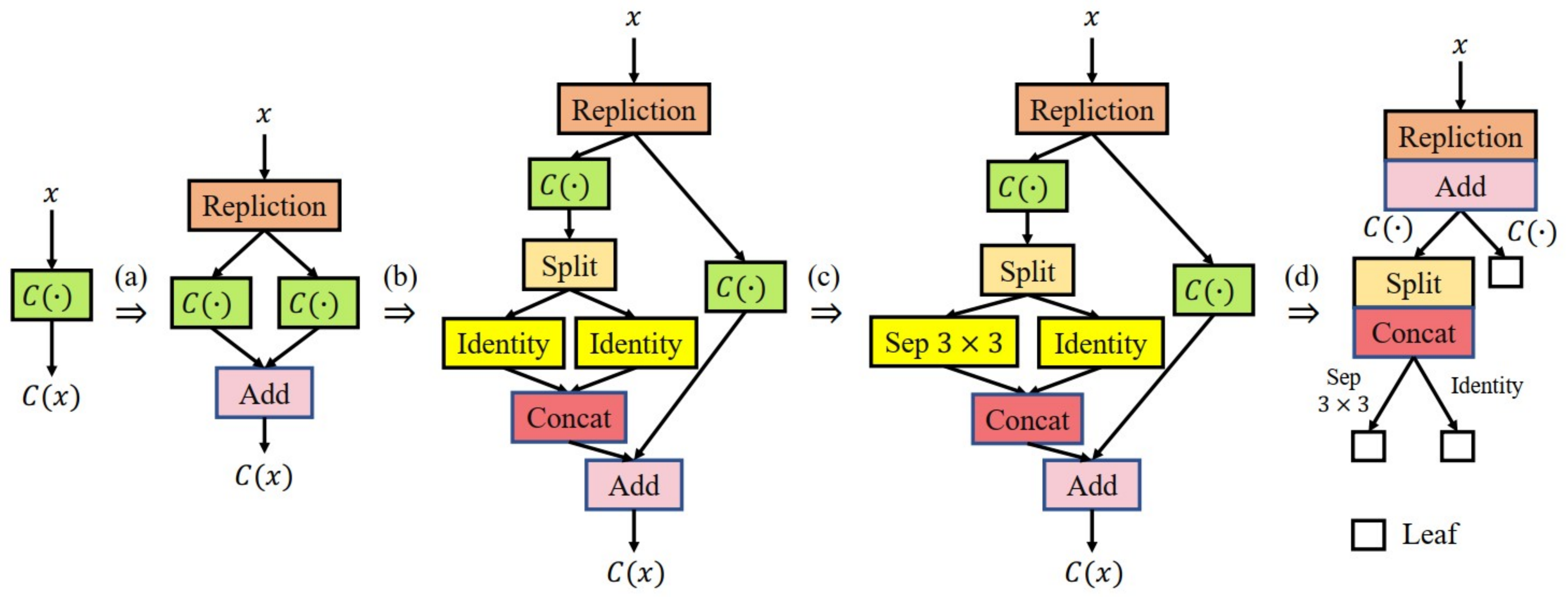}
	\caption{In Path-level EAS \cite{Path-level EAS}, an example of a process for implementing path-level network transformation using a multi-branch operation rather than a single layer. (a) Using the \textit{replication}-\textit{add} strategy to add branches to a single layer. (b) Using the \textit{split}-\textit{concat} strategy to further add branches to the network. (c) Replacing identity mapping with a $3\times 3$ depthwise-separable convolution. (d) The tree structure of the neural architecture in (c).}
	\label{fig:Path_Level}
\end{figure*}

Rather than widening or deepening the layers of the existing network in EAS \cite{EAS}, N2N learning \cite{N2N learning} compresses the teacher network by removing or shrinking the layers. In more detail, it compresses the teacher network through a two-stage operation selection: first, the layer removal is performed on the macro level, after which the layer shrinkage is performed on the micro-level. Reinforcement learning is used to explore the search space, while knowledge distillation \cite{Distilling the knowledge in a neural network} is used to train each generated neural architecture. In the next step, a locally optimal student network is learned. Using this method, under similar performance conditions, a compression ratio of more than $10\times$ is achieved for networks such as ResNet-34 \cite{Resnet}.  Moreover, unlike EAS \cite{EAS} and N2N learning \cite{N2N learning}, which can only deepen (remove) and widen (shrink) the network at the layer level, Path-level EAS \cite{Path-level EAS} realizes a network transformation at the path level. The inspiration behind this concept stems from the performance gains achieved by the multi-branch neural architecture included in the manually designed network \cite{Resnet,GoogLeNet,Inception-v4,Inception}, which achieves network path-level transformation by replacing a single layer with multi-branch operations incorporating allocation and merge strategies. Allocation strategies include: \textit{replication} and \textit{split}, while merge strategies include: \textit{add} and \textit{concatenation}. We present an example of the process of implementing a path-level network transformation by using a multi-branch operation rather than a single layer in Fig.\ref{fig:Path_Level}.  Another similar work, NASH-Net \cite{Simple and efficient architecture search for convolutional neural networks}, further proposes four network morphism types based on Net2Net \cite{Net2Net}. NASH-Net can begin with a pre-trained network, apply network morphism to generate a set of sub-networks, and obtain the best sub-network following a short period of training.  Then, after starting from the best sub-network, this process is iterated using the \textit{Neural Architecture Search by Hill-climbing} (NASH) to get the best neural architecture.

\begin{figure}[!tp]
	\centering
	\includegraphics[width=0.95\linewidth]{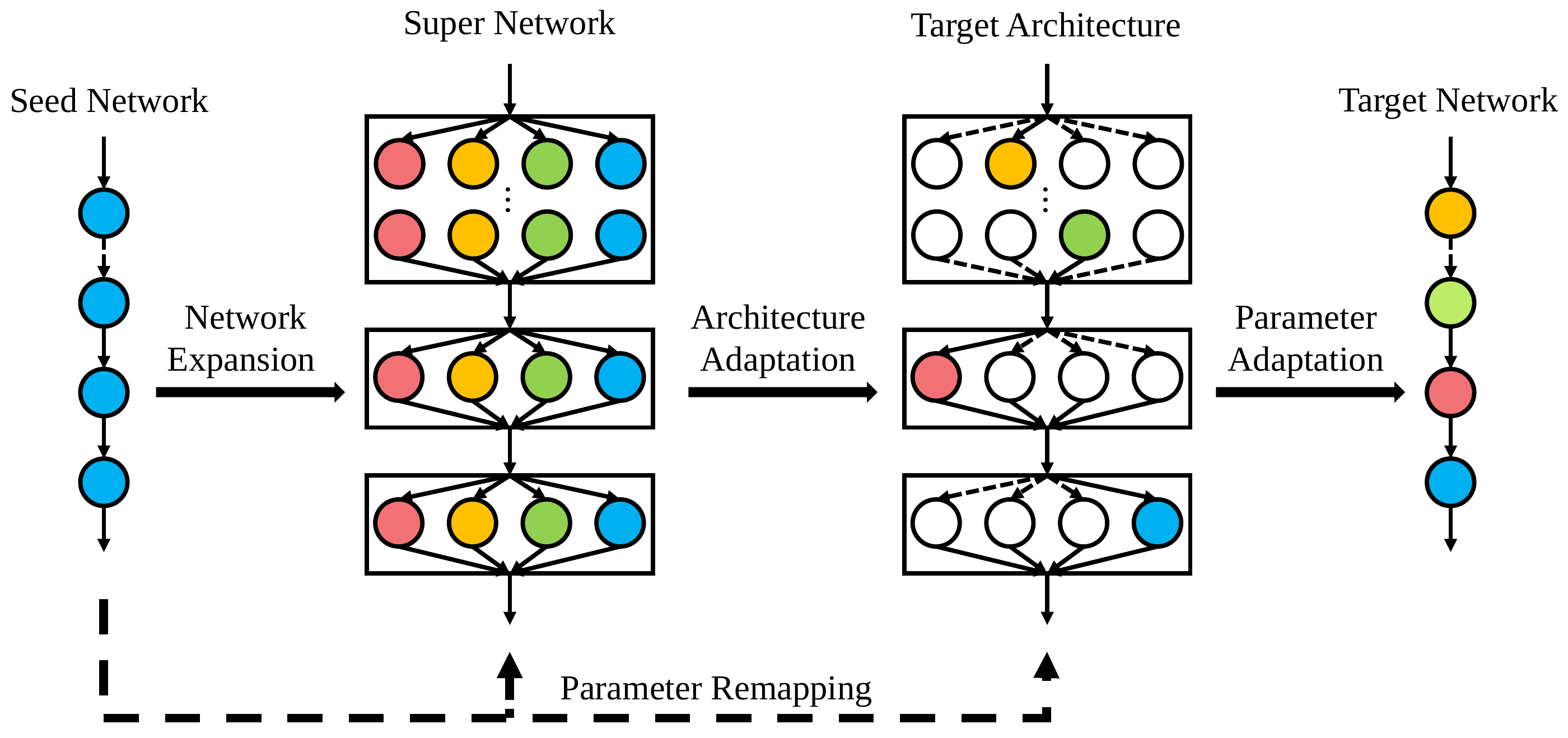}
	\caption{The overall framework of FNA \cite{FNA}. It starts from a seed network, expands this network into a super network in its operation set, and then uses the NAS method to adapt the neural architecture to obtain the target architecture. The seed network is then used to map the parameters to the super network and the target network to initialize the parameters. Finally, the target network is obtained via fine-tuning the target task.}
	\label{fig:FNA}
\end{figure}

For complex tasks such as semantic segmentation or object detection, previous works have often used networks designed for image classification, such as the backbone network. Under these circumstances, performance gains can be obtained by specifically designing networks for complex target tasks. Although some works \cite{Auto-deeplab,Customizable architecture search for semantic segmentation,Detnas} have used NAS to design backbone networks for semantic segmentation or object detection tasks, pre-training is still necessary and the computational cost is high. \textit{Fast Neural Network Adaptation} (FNA) \cite{FNA} proposes a method that can adapt a network's architecture and parameters to new tasks at almost zero cost. It starts from a seed network (a manually designed high-performance network), expands it into a super network in its operation set, and then uses the NAS method \cite{ENAS,DARTS,AmoebaNet-A} to adapt the neural architecture in a way that allows it to obtain the target architecture. Moreover, it uses the seed network to map the parameters to the super network and the target network to initialize the parameters. Finally, the target network is obtained by fine-tuning the target task. We illustrate this process in Fig.\ref{fig:FNA}. It is precisely due to the low cost of FNA in network transformation that NAS can design a special neural architecture for large-scale tasks, such as detection and segmentation. 

Different from the above methods which mainly focus on using the NAS method to improve the visual model, Evolved Transformer \cite{Evolved Transformer} is committed to using the NAS method to design a better feedforward architecture for seq2seq tasks. Specifically, Evolved Transformer first constructed a large search space and then run an evolutionary architecture search with warm starting by seeding our initial population with the Transformer, thereby search for better alternatives to the Transformer. Besides, to be able to dynamically allocate more computing resources to more promising candidate networks, Evolved Transformer also developed the Progressive Dynamic Hurdles method and achieved continuous improvement on four well-established language tasks.

In this section, we present a comprehensive overview of NAS based on architecture recycling. This method makes it possible to utilize a large number of high-performance networks that are previously artificially designed. This saves the NAS from having to search the neural architecture from scratch, which substantially reduces the number of unnecessary random searches required in the massive search space. Compared with other optimization strategies, there are relatively few studies on NAS based on neural architecture recycling.

\subsection{Incomplete Training}
\label{sec:Incomplete Training}
The key technology behind NAS involves using a search strategy to find the best neural architecture by comparing the performance of a large number of candidate neural architectures. Accordingly, the performance ranking of candidate neural architectures is extremely important. Early versions of NAS \cite{Neural architecture search with reinforcement learning,MetaQNN,Large-scale Evolution,GeNet} usually fully train the candidate neural architecture, then obtain the rankings of the candidate neural architectures based on their performance on the validation set. However, this method is excessively time-consuming because there are excessively many candidate neural architectures to compare.

It should however be noted here that these works have also used some methods to accelerate the ranking of candidate neural architectures. For example, NAS-RL \cite{Neural architecture search with reinforcement learning} uses parallel and asynchronous updates \cite{Large scale distributed deep networks} to accelerate candidate neural architecture training. MetaQNN \cite{MetaQNN} compares the performance of the candidate neural architecture after the first epoch of training with the performance of the random predictor to determine whether it is necessary to reduce the learning rate and restart training. Large-scale Evolution \cite{Large-scale Evolution} allows the mutated child neural architecture to inherit the weight of the parent to the greatest possible extent, thereby reducing the burden associated with retraining the candidate neural architecture. However, there are still a large number of child networks whose structural changes are unable to inherit the weight of their parents after mutation, meaning that these candidate networks will be forced to retrain. Although the above methods also accelerate the training of candidate neural architectures to a certain extent, they still require a lot of computing power and their acceleration effects are relatively limited. Therefore, it is necessary to conduct some research that can be used to further accelerate the training of candidate neural architectures to obtain their relative ranking.

\subsubsection{Training from scratch?}
Can we only train each candidate's neural architecture from scratch? This may ignore the interconnection between neural architectures. Parameter sharing brings us a new possibility.

When treating the candidate neural architecture as an independent individual, each candidate neural architecture is trained from scratch, after which the candidate neural architecture is ranked according to their performance on the validation set. This may provide a more accurate ranking, as has occurred in other works \cite{Neural architecture search with reinforcement learning, NASNet, PNAS, Hierarchical-EAS}. In this process, the parameters of each trained candidate neural architecture are directly discarded. This does not result in full utilization of these trained parameters; accordingly, a new idea of parameter sharing has emerged.

\begin{figure}[!tp]
	\centering
	\includegraphics[width=0.75\linewidth]{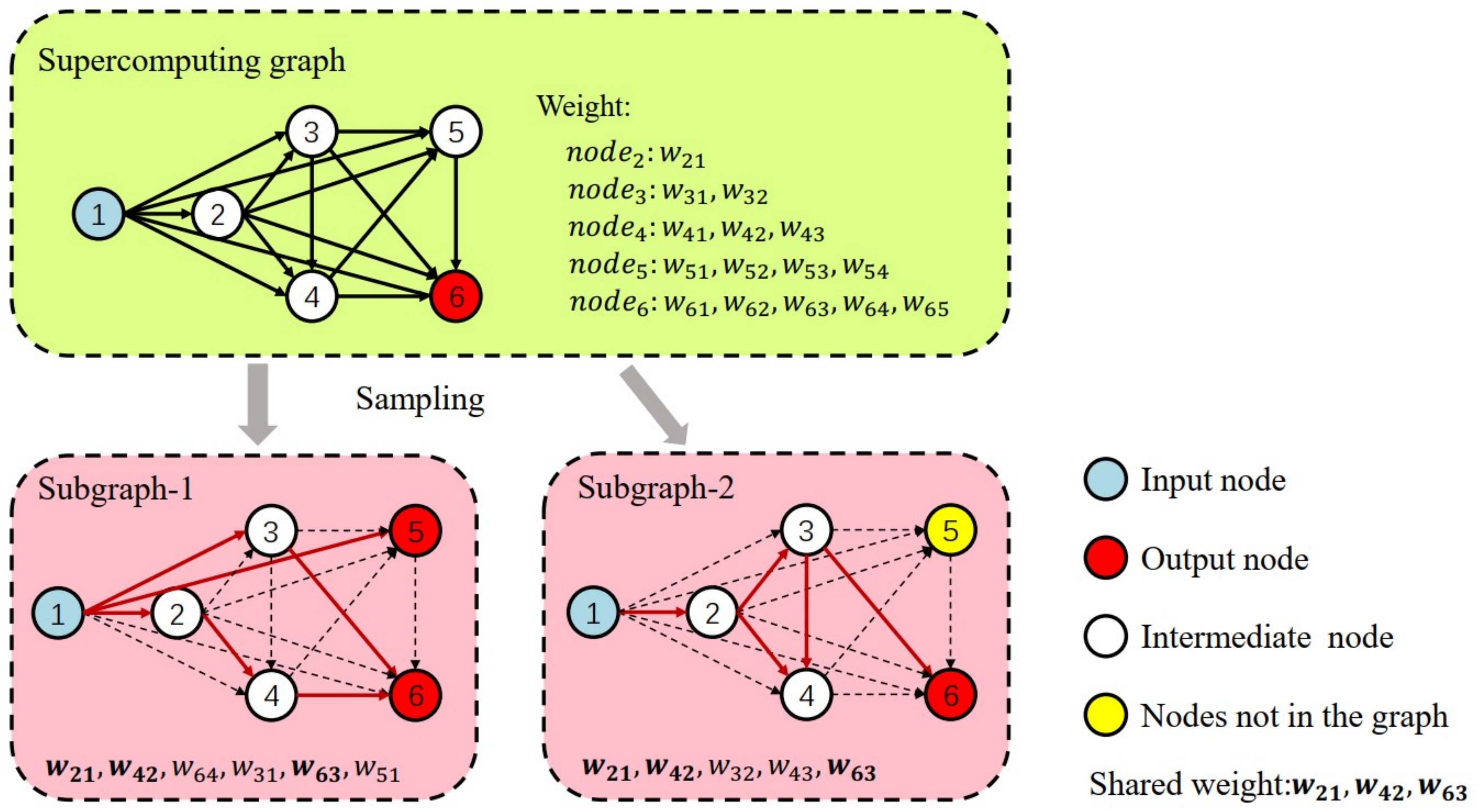}
	\caption{Parameter sharing mechanism in ENAS \cite{ENAS}. The candidate neural architecture in NAS can be represented as a directed acyclic subgraph that is sampled in a supercomputing graph constructed by the search space. The nodes in the graph represent local calculations, while the edges represent information flow. Each node has its corresponding weight parameter, as indicated in the upper right. Only when a specific edge is sampled will the corresponding parameter be activated and updated. The ENAS mechanism allows all subgraphs (that is, candidate neural architectures) to share parameters.}
	\label{fig:ENAS}
\end{figure}

ENAS \cite{ENAS} is the first NAS work to explicitly proposes parameter sharing. The ENAS work has noted that the candidate neural architecture in NAS can be regarded as a directed acyclic subgraph, which is in a supercomputing graph constructed by the search space. We illustrate this sampling process in Fig.\ref{fig:ENAS}. Based on this observation, ENAS uses LSTM as a controller for use in searching the optimal subgraph on a large computation graph to obtain the neural architecture. In transfer learning and multi-task learning, the weights obtained by training a model designed for a specific task on a data set are also applicable to other models designed for other tasks \cite{CNN features off-the-shelf,Transfer learning for low-resource neural machine translation,Multi-task sequence to sequence learning}. Encouraged by this, ENAS proposes forcing the sharing of parameters among all different child models (candidate architecture). Through the use of this mechanism, the child models can obtain empirical performance, thereby avoiding the need to completely train each child model from scratch. We present an example of different subgraphs sharing weights in Fig.\ref{fig:ENAS}. The supercomputing graph can be expressed as a DAG: the nodes in the graph are defined as local calculations, while the edges represent the flow of information. Each node has its corresponding weight parameter, as illustrated in the upper right of Fig.\ref{fig:ENAS}. However, the corresponding parameters can only be activated when a specific edge is sampled. The ENAS mechanism allows all subgraphs (i.e. candidate neural architectures) to share parameters. Therefore, EANS has greatly improved search efficiency compared to \cite{Neural architecture search with reinforcement learning,NASNet,PNAS,Hierarchical-EAS}.

Subsequently, CAS \cite{CAS} explored a multi-task architecture search based on ENAS. This approach extends NAS to transfer learning across data sources, and further introduces a novel continuous architecture search to solve this 'forgetting' problem in the continuous learning process. This enables CAS to inherit the experience gained from the previous task when training a new task, thereby allowing the model parameters can to continuously trained. This is highly beneficial to NAS research on multi-tasking. Moreover, AutoGAN \cite{Autogan} first introduced NAS into \textit{generative adversarial networks} (GANs) \cite{GAN} and used the \textit{Inception score} (IS) \cite{IS} as the reward value of RL to accelerate the search process through parameter sharing ENAS \cite{ENAS} and dynamic-resetting. Further, progressive GAN training \cite{Progressive growing of gans for improved quality} has been used to introduce \textit{multi-level architecture search} (MLAS) into AutoGAN and gradually implement NAS. Compared with the most advanced manual GANs \cite{Progressive growing of gans for improved quality,Dist-gan,mmd-gan,MGAN}, AutoGAN is highly competitive. The parameter sharing mechanism is also used to accelerate the deployment research of the NAS architecture model across multiple devices and multiple constrained environments. At the kernel level, OFA \cite{Once for all} uses an elastic kernel mechanism to meet the application needs of multi-platform deployment and the diverse visual needs of different platforms. Small kernels share the weight of large kernels; this is done to avoid repeatedly centering sub-kernels (centering sub-kernels is used for both an independent kernel and a part of a large kernel) to reduce the performance of certain sub-networks, OFA also introduces a kernel transformation matrix. At the network level, OFA recommends training the largest network first, while the smaller network shares the weight of the larger network before fine-tuning. Moreover, the weight of the large network can provide a small network with good initialization, which greatly accelerates the training efficiency.

Besides, the one-shot based method also employs parameter sharing. SMASH \cite{SMASH} proposes to train an auxiliary HyperNet \cite{HyperNet}, which is used to generate weights for other candidate neural architectures. Besides, SMASH also uses the early training performance of different networks derived from the research in Hyperband \cite{Hyperband} to provide meaningful guidance suggestions for the ranking of candidate neural architectures. Parameter sharing is primarily reflected in the hypernetwork and between candidate neural architectures. The use of the auxiliary HyperNet avoids the need to completely train each candidate neural architecture. By comparing the performance of the candidate neural architectures using weights generated by the HyperNet on the verification set, their relative rankings can be obtained; this allows SMASH to quickly obtain the optimal neural architecture at the cost of a single training session. Understanding One-Shot Models \cite{Understanding One-Shot Models} conducts a comprehensive analysis of the rationality of the parameter sharing approaches used in SMASH \cite{SMASH} and ENAS \cite{ENAS}. Besides, Understanding One-Shot Models also discussed the necessity of the hypernetwork in SMASH and the RL controller in ENAS, pointing out that a good enough result can be obtained without the use of the hypernetwork and RL controller. Unlike SMASH, which encodes an architecture into a three-dimensional tensor via a memory channel scheme, Graph HyperNetwork (GHN) \cite{GHN} employs computation graphs to represent the neural architecture, then uses graph neural networks to perform architecture searches. Compared to SMASH, which can only use the hypernetwork to predict some weights, GHN can predict all the free weights through the use of a graph model. Therefore, network topology modeling-based GHN can predict network performance faster and more accurately than SMASH. 

A typical one-shot NAS must randomly sample a large number of candidate architectures from the hypernetwork using parameter sharing, then evaluate these architectures to find the best one \cite{SMASH,Understanding One-Shot Models}. SETN \cite{SETN} noted that finding the best architecture from these sampled candidate architectures is extremely difficult. This is because, in the relevant NAS \cite{SMASH,Understanding One-Shot Models,GDAS}, the shared parameters are closely coupled with the learnable architectural parameters. This will introduce deviations into the template parameters, which will cause some of the learnable architectural parameters to be more biased towards simple networks (these networks have fewer layers and are more lightweight). Because they converge faster than more complex networks, this will lead to a simplified search architecture. At the same time, it also results in the sampled candidate architectures having a very low good rate. To this end, SETN adopts a uniformly stochastic training strategy to treat each candidate architecture fairly, meaning that they are fully trained to obtain more accurate verification performance. Besides, SETN is also equipped with a template architecture estimator. Unlike the random sampling methods previously used in Understanding One-Shot Models \cite{Understanding One-Shot Models}  and SMASH \cite{SMASH}, the estimator in SETN can be used to determine the probability that the candidate architecture has a lower verification loss, as well as to ensure that the low verification loss architecture with a higher probability will be selected for one-shot evaluation. At the same time, the estimator is trained on the validation set. Therefore, SETN improves the excellent rate of the sampling candidate architecture compared to Understanding One-Shot Models \cite{Understanding One-Shot Models} and SMASH \cite{SMASH}, making it more likely to find the optimal architecture.

\cite{Evaluating the search phase of neural architecture search}, through evaluating the effectiveness of the NAS search strategy, find that the weight sharing strategy in ENAS \cite{ENAS} resulted in inaccurate performance evaluation of the candidate architecture, making it difficult for the NAS to identify the best architecture. Besides, the research of Fairnas \cite{Fairnas} and \cite{PC-NAS} also demonstrates that candidate neural architectures based on these parameter sharing methods also cannot be adequately trained, which lead to an inaccurate ranking of candidate neural architectures. In NAS works based on gradient optimization \cite{DARTS,ProxylessNAS,Fbnet}, the joint optimization of supernet weights and architectural parameters also introduces bias between sub-models. In light of this, DNA \cite{Blockwisely Supervised Neural Architecture Search with Knowledge Distillation} proposes to modularize the NAS's large-scale search space to ensure that the candidate architecture is adequately trained to reduce the representation shift caused by parameter sharing. Besides, DNA \cite{Blockwisely Supervised Neural Architecture Search with Knowledge Distillation} also uses block-wise search to evaluate all candidate architectures within the block. These methods are used to evaluate candidate architectures more accurately. GDAS-NSAS \cite{Overcoming Multi-Model Forgetting in One-Shot NAS with Diversity Maximization} also considered and improved the weight sharing mechanism in one-shot NAS, proposing an NSAS loss function to solve the problem of multi-model forgetting (when weight sharing is used to sequentially train a new neural architecture, the performance of the previous neural architecture is reduced) that arises due to weight sharing during the super network training process. Finally, GDAS-NSAS \cite{Overcoming Multi-Model Forgetting in One-Shot NAS with Diversity Maximization} applies the proposed method to RandomNAS \cite{RandomNAS} and GDAS \cite{GDAS}; this approach effectively suppresses the multi-model forgetting problem and consequently improves the training quality of the supernet,

Differentiable neural architecture search also employs similar parameter sharing ideas. Examples include DARTS-like work \cite{DARTS,I-DARTS,P-DARTS,PC-DARTS}; for details, refer to Section.\ref{sec:Continuous Search Strategy}. In ENAS, a controller is used to sample subgraphs in a supercomputing graph. Subgraphs with the same information flow share parameters in the search phrase, so only the sampled subgraphs need to be optimized in each iteration. The difference is that the DARTS-like method chooses to optimize a super network directly, and the best sub-network is decoupled from the super network according to the learned mixed operation weights. Parameters are shared among different sub-networks in the super network. Moreover, optimization strategies based on neural architecture recycling can often be initialized with the help of function preservation \cite{Net2Net} to inherit the parameters of the template network, thereby avoiding the retraining of sub-neural architecture. More detailed content can be found in Section.\ref{sec:Neural Architecture Recycle}, for example: EAS \cite{EAS}, Path-level EAS \cite{Path-level EAS} and N2N learning \cite{N2N learning}, etc.

\subsubsection{Training to convergence?}
Can we only train each candidate neural architecture to convergence? This may ignore the guiding role of early performance curves in predicting the potential of neural architecture. Early termination brings us a new possibility.

To quickly analyze the effectiveness of a specific model in deep learning, human experts often judge whether this model is necessary to continue training according to the model's learning curve. Therefore, the training of models with no potential will be terminated as soon as possible; this means that there is no need to wait for this model to converge, which saves resources for new exploration. Similar strategies can also be used to rank the performance of NAS candidate architectures: for candidate architectures with no potential, training can be terminated early, while more adequate training can be obtained for more promising architectures.

Early termination of training is not a new idea; many researchers have done a large amount of related research on this subject. For example, \cite{Speeding up automatic hyperparameter optimization of deep neural networks by extrapolation of learning curves} uses the probabilistic method to simulate the learning curve of the deep neural network and terminates the training of poorly running models in advance. However, this approach requires a long early training period to accurately simulate and predict the learning curve. \cite{Learning curve prediction with Bayesian neural networks} extends \cite{Speeding up automatic hyperparameter optimization of deep neural networks by extrapolation of learning curves}; in \cite{Learning curve prediction with Bayesian neural networks}, the probability model of the learning curve can be set across hyperparameters, and a mature learning curve is used to improve the performance of the Bayesian neural network. Similar strategies have also been used to solve hyperparameter optimization problems \cite{Hyperband,Speeding up hyper-parameter optimization by extrapolation of learning curves using previous builds}.

\begin{figure}[!tp]
	\centering
	\includegraphics[width=0.5\linewidth]{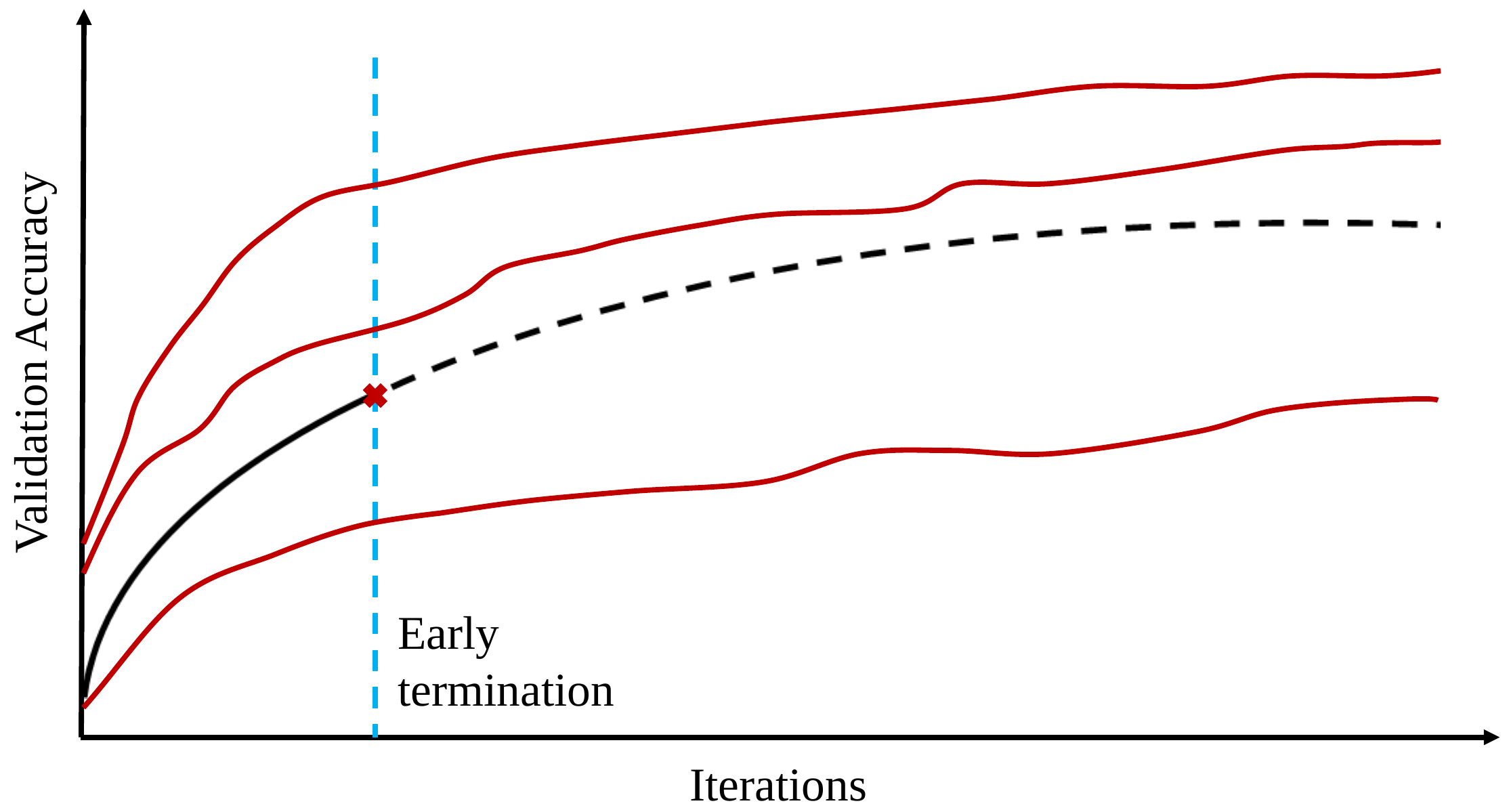}
	\caption{Assuming learning curve for performance prediction. The black curve represents an attempt to use an observable early performance learning curve (solid line) to predict the neural architecture's final performance on the validation set \cite{PNAS,Peephole,Accelerating neural architecture search using performance prediction}. The three red curves indicate the hypothesis that the performance of the neural architecture during early training and convergence is consistent with ranking \cite{MdeNAS}. Using performance prediction and early performance ranking assumptions can facilitate the termination of the training of neural architectures that lack potential as early as human experts, as well as the use of minimal computing resources to explore more potential neural architectures.  This speeds up the NAS architecture search process.}
	\label{fig:EarlyTermination}
\end{figure}

The above methods are based on the use of partially observed early performance to predict the learning curve and design the corresponding machine learning model. To imitate human experts to the extent that NAS search can also automatically identify below-standard candidate architectures and terminate their training early, \cite{Accelerating neural architecture search using performance prediction} combines learning curve prediction with NAS tasks for the first time. This approach builds a set of standard frequentist regression models and obtains the corresponding simple features from the neural architecture, hyperparameters, and early learning curve. These features are then used to train the frequentist regression model and are then used to predict the final verification set performance of the neural architecture with early training experience. Performance prediction is also used in PNAS \cite{PNAS}. To avoid the need to train and evaluate all child networks, it learns a predictor function that can be trained based on the observable early performance of the cell. The predictor is used to evaluate all candidate cells, then the top-$k$ cells are selected, and this process is repeated until a sufficient number of blocks of cells are found. The black curve in Fig.\ref{fig:EarlyTermination} shows the learning curve prediction trying to predict the final performance from the premature learning curve.

NAO \cite{NAO} uses a performance predictor similar to previous work \cite{PNAS,Peephole,Accelerating neural architecture search using performance prediction}. This is dissimilar to PNAS \cite{PNAS}, which uses a performance predictor to evaluate and select the generated neural architecture to speed up the search process. In NAO \cite{NAO}, after the encoder completes the continuous representation of the neural architecture, the performance predictor is taken as the optimization goal of the gradient ascent. By maximizing the output of the performance predictor $f$, the continuous representation of the best neural architecture can be obtained. Finally, the decoder is used to get the final discrete neural architecture. Unlike previous NAS based on performance prediction \cite{PNAS,Accelerating neural architecture search using performance prediction,Peephole}, in multinomial distribution learning for NAS, MdeNAS \cite{MdeNAS} put forward a performance ranking hypothesis: that is, the relative performance ranking of the neural architecture at each training stage is consistent.  In other words, the neural architecture that performed well in the early days still maintains good performance when the training converges. MdeNAS \cite{MdeNAS} has conducted a  large number of experiments to verify this hypothesis; according to these results, the early performance of candidate architectures can be used to quickly and easily obtain their relative performance rankings, thereby speeding up the neural architecture search process. We illustrate an early performance ranking hypothesis in Fig.\ref{fig:EarlyTermination} (as shown by the three red curves).

In this section, we focus on the challenge of fully trained candidate architectures, starting with the necessity of two aspects (training from scratch and training to convergence), and comprehensively and systematically summarizing the existing work. Compared with other optimization strategies, this aspect of the research work is relatively small, but still very necessary.

\section{Performance Comparison}
\label{sec:Performance comparison}
\begin{figure}[!tp] 
	\centering    %居中
	\subfloat[The general framework of RL  in the context of NAS.] %第一张子图
	{\centering          %子图居中
		\includegraphics[width=0.37\linewidth]{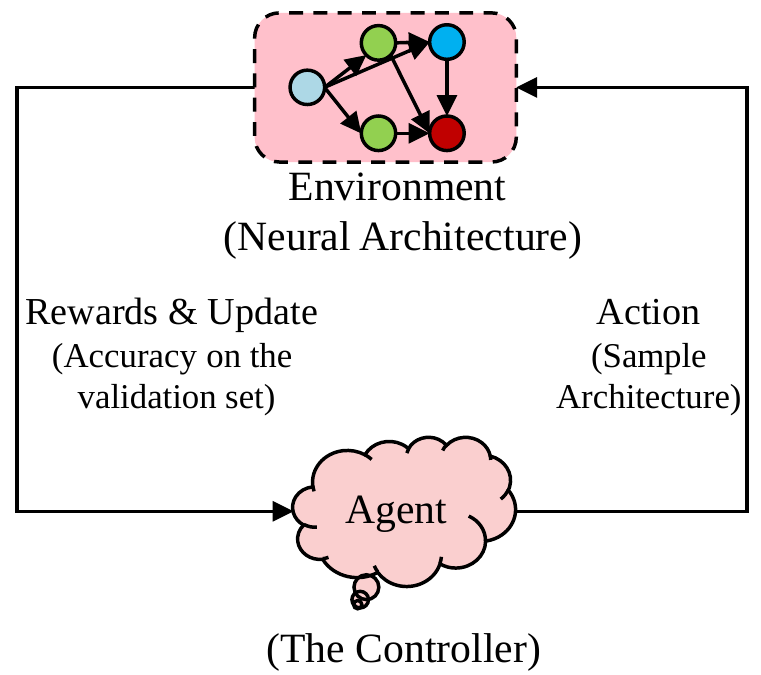}
		\label{fig:RL}
	}\hspace{4mm}
	\subfloat[The general framework of SMBO in the context of NAS.] %第二张子图
	{\centering      %子图居中
		\includegraphics[width=0.56\linewidth]{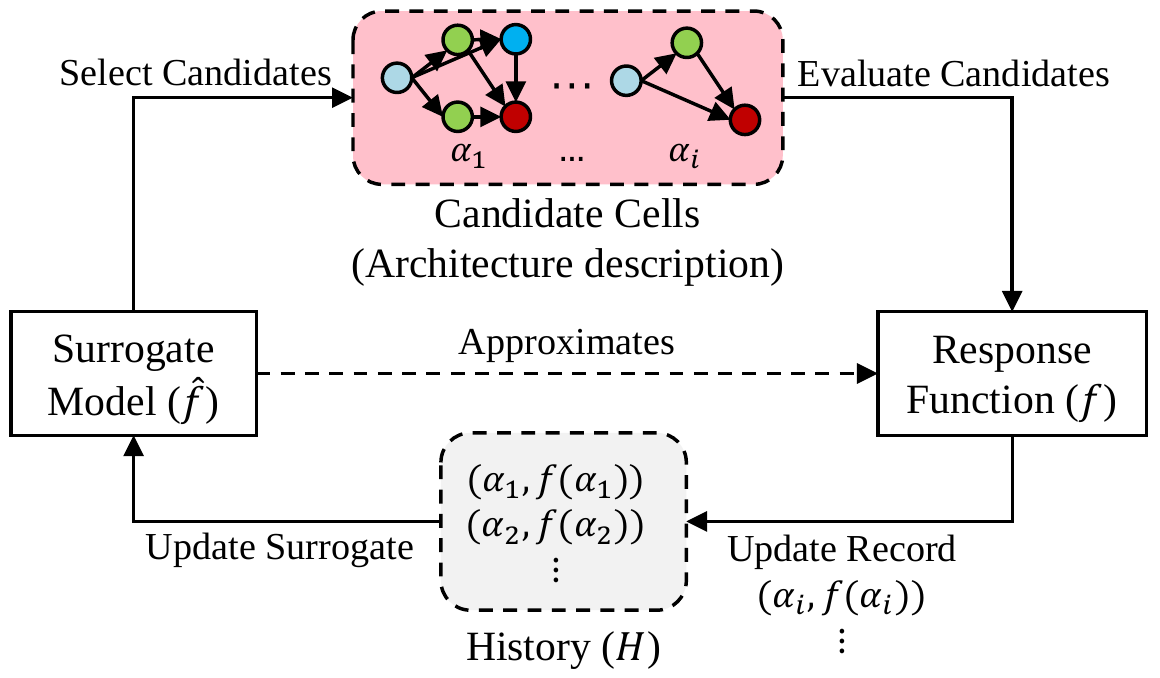}
		\label{fig:SMBO}
	}\\
	\subfloat[The general framework of EA in the context of NAS.] %第二张子图
	{\centering      %子图居中
		\includegraphics[width=0.8\linewidth]{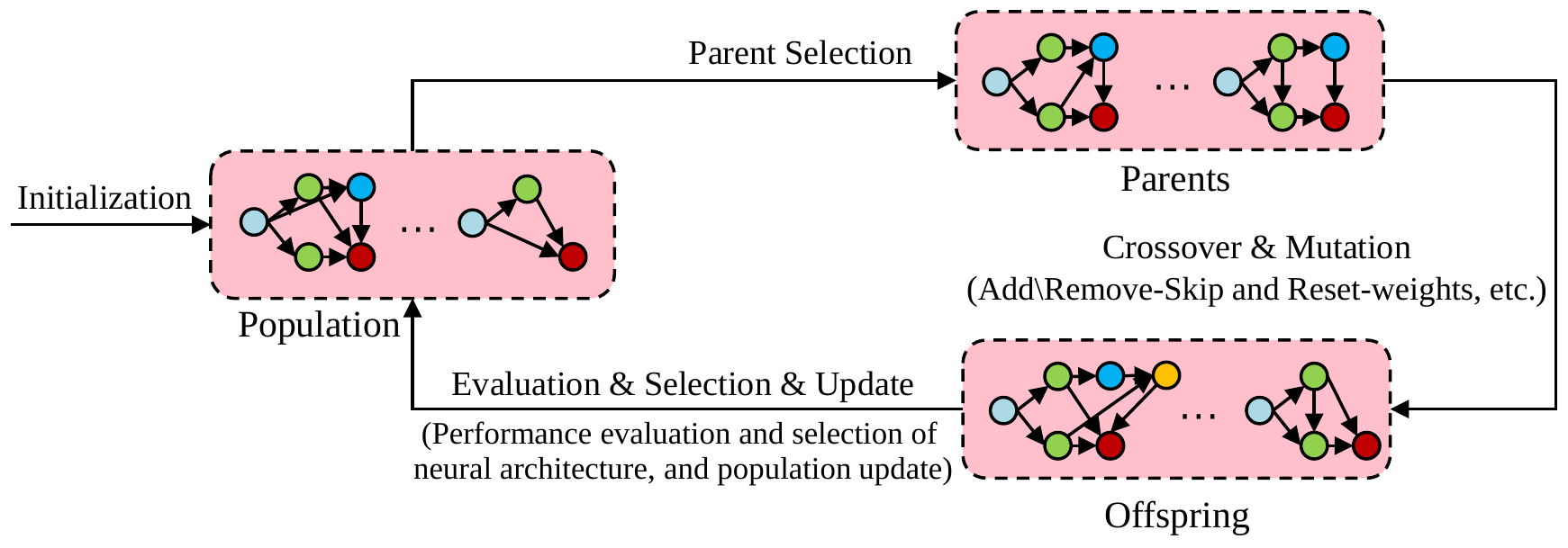}
		\label{fig:EA}
	}
	\caption{Comparison of RL, SMBO, and EA general frameworks in the context of NAS. }
	\label{fig:RL_SMBO_EA}  %图片引用标记

\end{figure}

NAS is a highly promising study. 
% In Section \ref{sec:Characteristics of early NAS}, we analyzed the common characteristics of early NAS and summarized the challenges that prevented early NAS from being widely used. In Section \ref{sec:Optimization Strategy}, we conduct a comprehensive and systematic review of the solutions adopted by our existing work to address these challenges. 
In this section, we classify and compare the performance of existing NAS based on mainstream search methods \cite{A survey on neural architecture search,Neural Architecture Search: A Survey}, while also reporting the optimization strategies they use according to Section  \ref{sec:Optimization Strategy}. These search methods primarily include the following: \textit{reinforcement learning} (RL), \textit{evolutionary algorithm} (EA), \textit{gradient optimization} (GO), \textit{random search} (RS) and \textit{sequential model-based optimization} (SMBO) \cite{SMBO}. We summarized in Fig.\ref{fig:RL_SMBO_EA} the general framework comparison of RL, SMBO, and EA in the context of NAS. 
Fig.\ref{fig:RL} shows the general framework of RL in the context of NAS. The agent (controller) makes corresponding actions (sampling neural framework) to modify the system state (neural architecture), and the environment gives the agent corresponding rewards (accuracy on the verification set) and updates the agent accordingly (adjusting algorithm parameters). 
Fig.\ref{fig:SMBO} shows the general framework of SMBO in the context of NAS. The surrogate model $\hat{f}$ (performance predictor) evaluates all candidate cells $\alpha$ (the structure description of the neural architecture) \cite{PNAS} and selects promising candidate cells, and then evaluates the performance of the neural architecture composed of the candidate cells on the validation set and get the response value $f(\alpha)$ (loss value) of the corresponding response function $f$ \cite{ResponseFunction}. The cell $\alpha_i$ and the corresponding response value $f(\alpha_i)$ are added as a new meta instance $(\alpha_i, f(\alpha_i))$ to the history record $H$ of cell measurement performance, and the surrogate model is updated again according to $H$ until convergence. Through iteration, the surrogate model value $\hat{f}(\alpha)$ of the cell $\alpha$ is continuously approximated to the corresponding response value $f(\alpha)$ to avoid time-consuming training steps. 
Fig.\ref{fig:EA} shows the general framework of EA in the context of NAS. The algorithm first initializes the population and selects parents, then cross or mutate based on the parent to generate a new individual offspring (neural architecture), and evaluate the adaptability of the offspring (the performance evaluation of the new neural architecture), and select a group of individuals with the strongest adaptability (the best performing neural architecture) to update the population. 
GO has been described in detail in Fig.\ref{fig:DARTS}, so it is no longer repeated here.

We intend to obtain their similarities and differences from these summaries. The situation is not as simple as it might appear.  In fact, it is relatively difficult to compare NAS performance, because NAS lacks certain baselines. 
Besides, different types of NAS have great differences in preprocessing, hyperparameters, search space, and trick, which increase the difficulty of NAS performance comparison. 
These tricks include learning rate decay, regularization (e.g., DropPath \cite{NASNet,Deep networks with stochastic depth}), augmenting techniques (e.g., Cutout \cite{cutout}), etc. The random search strategy is considered to be a strong baseline. For example, in \cite{Evaluating the search phase of neural architecture search}, the random search identifies the best RNN cells compared to other strategies. The findings presented in \cite{NAS evaluation is frustratingly hard}, \cite{Hierarchical-EAS} and \cite{RandomNAS}  also prove this. Therefore, \cite{NAS evaluation is frustratingly hard} uses the average architecture of the random search strategy as the baseline for comparison. 

\begin{table}[!tp]
	\centering
	\caption{The performance comparison between the state-of-the-art NAS and mainstream artificial networks on CIFAR-10. In the interests of clarity, we classify NAS according to mainstream search methods: reinforcement learning (RL), evolutionary algorithm (EA), gradient optimization (GO), random search (RS) and sequential model-based optimization (SMBO). At the same time, the optimization strategy used in NAS is reported based on the analysis Section \ref{sec:Optimization Strategy}. Cutout is an augmentation technology used in \cite{cutout}.}
	\label{tab:CIFAR-10}
    \tiny
    \resizebox{\textwidth}{80mm}{
	\begin{tabular}{|c|l|c|cccc|c|c|c|}
		\hline 
		\multirow{4}{*}{\makecell[c]{Search\\ method}} & \multirow{4}{*}{Reference}&\multirow{4}{*}{Venue} & \multicolumn{4}{c|}{Optimization Strategy} & \multirow{4}{*}{\makecell[c]{Error\\Acc (\%)}} & \multirow{4}{*}{\makecell[c]{Params \\(Millions)}} &\multirow{4}{*}{\makecell[c]{GPU\\Days}} \\ 
		&&&\makecell[c]{Modular \\ search\\space} &\makecell[c]{Continuous \\search \\strategy} & \makecell[c]{Architecture\\ recycle} & \makecell[c]{Incomplete\\ training}&&&\\
		\hline \hline 
		
		%Human
		\multirow{6}{*}{Human} & WRN \cite{WRN} &CVPR16& && &  & 3.87& 36.2&-  \\  
		&Shark \cite{Shark} &CoRR17 & && &  &3.55& 2.9 &-  \\ 
		&\makecell[l]{PyramidSepDrop \cite{PyramidSepDrop}} &CoRR16 & && &  &  2.67 &26.2& -  \\ 
		&ResNet \cite{Deep networks with stochastic depth}&ECCV16  & && &  &6.41& 1.7 &-  \\ 
		&Fractalnet \cite{Fractalnet}&ICLR17  & && &  & 5.22 &38.6 &-\\
		&DenseNet-BC \cite{DenseNet}&CVPR17 & && &  & 3.46& 25.6 &-\\
		\hline\hline 
		
		%RL
		\multirow{11}{*}{RL} &NAS-RL \cite{Neural architecture search with reinforcement learning}&ICLR17  &  &  &  &  & 3.65 & 37.4 & 22,400\\ 
		& MetaQNN \cite{MetaQNN} &ICLR17 & & &  &  & 6.92 & 11.2 & 100 \\ 
		& EAS \cite{EAS}&AAAI18  &  & & \checkmark & \checkmark  & 4.23 & 23.4 & 10 \\ 
		& NASNet-A \cite{NASNet}&CVPR18  &\checkmark &  & & & 3.41 &3.3 &2,000 \\
		& \makecell[l]{NASNet-A + Cutout \cite{NASNet}} &CVPR18 &\checkmark &  & & &2.65 &3.3 &2,000 \\
		& Block-QNN \cite{Block-QNN}&CVPR18  &\checkmark &  & & &3.54   &39.8 &96 \\
		& Path-level EAS \cite{Path-level EAS}&ICML18   & &  &\checkmark &\checkmark &2.99 &5.7 &200 \\
		&\makecell[l]{Path-level EAS $+$ Cutout \cite{Path-level EAS}}&ICML18  & &  &\checkmark & \checkmark&2.49 &5.7 &200 \\
		&N2N learning  \cite{N2N learning} &ICLR18 & &&\checkmark & \checkmark & 6.46 & 3.9 &2.1  \\ 
		&\makecell[l]{ProxylessNAS-R $+$ Cutout \cite{ProxylessNAS}}  &ICLR19& &  & & \checkmark  &2.30 &5.8& N/A  \\ 
		&FPNAS $+$ Cutout   \cite{FPNAS} &ICCV19 &\checkmark && & \checkmark &3.01 & 5.8 &0.8  \\ 
		
		%EA
		\hline\hline 
		\multirow{8}{*}{EA} &\makecell[l]{Large-scale Evolution \cite{Large-scale Evolution}} &ICML17 & && &  \checkmark&5.40& 5.4 &2,600  \\ 
		&GeNet \cite{GeNet}&ICCV17  & && &  & 5.39& N/A &17\\ 
		&\makecell[l]{\makecell[l]{Genetic Programming CNN \cite{A genetic programming approach to designing convolutional neural network architectures} }}&GECC17 & && &  & 5.98& 1.7& 14.9  \\ 
		&Hierarchical-EAS \cite{Hierarchical-EAS}&ICLR18  &\checkmark && &  &3.75 &15.7& 300 \\ 
		
		&NASH-Net  \cite{Simple and efficient architecture search for convolutional neural networks} &ICLR18 & &&\checkmark & \checkmark &5.20 &19.7& 1 \\ 
		&\makecell[l]{Neuro-Cell-based Evolution \cite{Deep learning architecture search by neuro-cell-based evolution with function-preserving mutations}}&\makecell[c]{ECML-KDD18}&\checkmark &&\checkmark &\checkmark  &3.57 &5.8& 0.5 \\ 
		&AmoebaNet-A \cite{AmoebaNet-A}&AAAI19  &\checkmark && &  &3.34 &3.2 &3,150 \\ 
		&Single-Path One-Shot NAS \cite{Single path one-shot neural architecture search with uniform sampling}&CoRR19  &\checkmark && &  \checkmark & N/A & N/A & N/A\\ 
		\hline\hline 
		
		%GO
		\multirow{24}{*}{GO} 
		& \makecell[l]{ENAS $+$ micro \cite{ENAS}}  &ICML18&\checkmark &\checkmark& & \checkmark & 3.54 & 4.6 &0.5  \\ 
		& \makecell[l]{ENAS $+$ micro $+$ Cutout \cite{ENAS}}  &ICML18&\checkmark &\checkmark& & \checkmark & 3.54 & 4.6 &0.5  \\ 
		& ENAS $+$ macro \cite{ENAS} &ICML18&&\checkmark& & \checkmark &4.23 & 21.3 &0.32  \\ 
		
		& SMASH \cite{SMASH}&ICLR18  & &\checkmark& &   \checkmark&4.03  &16  &1.5  \\ 
		&\makecell[l]{Understanding One-Shot Models \cite{Understanding One-Shot Models}}&ICML18 & \checkmark & \checkmark& &   \checkmark&4.00 &5.0& N/A  \\ 
		&\makecell[l]{Maskconnect \cite{Maskconnect}}&ECCV18 & \checkmark & \checkmark & &   \checkmark &3.27 & N/A & N/A  \\ 
		&\makecell[l]{DARTS  ($1^{st}$ order) $+$ Cutout \cite{DARTS}}&ICLR19  & \checkmark & \checkmark & & \checkmark &3.00  &3.3  & 1.5 \\ 
		&\makecell[l]{DARTS  ($2^{nd}$ order) $+$ Cutout \cite{DARTS}}&ICLR19  & \checkmark & \checkmark & & \checkmark &2.76  &3.3  & 4 \\
		&SNAS $+$ Cutout  \cite{SNAS}&ICLR19  & \checkmark & \checkmark& &  \checkmark & 2.85 &2.8& 1.5  \\ 
		&GHN \cite{GHN} &ICLR19 & \checkmark &\checkmark& &  \checkmark & 2.84& 5.7 &0.84  \\ 
		&\makecell[l]{ProxylessNAS-G $+$ Cutout \cite{ProxylessNAS}}  &ICLR19& & \checkmark & & \checkmark  &2.08 &5.7& N/A  \\ 
		&PARSEC $+$ Cutout \cite{PARSEC} &CoRR19 & \checkmark & \checkmark& &  \checkmark & 2.81& 3.7 &1   \\ 
		&BayesNAS \cite{BayesNAS} &ICML19 &\checkmark &\checkmark & &  \checkmark&  2.81&3.4  & 0.2 \\ 
		& P-DARTS $+$ Cutout \cite{P-DARTS} &ICCV19& \checkmark&\checkmark& &\checkmark  &  2.50& 3.4 & 0.3 \\
		&SETN \cite{SETN}  & ICCV19 & \checkmark & \checkmark & & \checkmark  & 2.69 & N/A & 1.8 \\ 
		&DATA $+$ Cutout \cite{DATA}  &NeurIPS19& \checkmark&\checkmark& &\checkmark  & 2.59 &3.4 & 1\\ 
		&TAS \cite{TAS}  & NeurIPS19 & &\checkmark&\checkmark & \checkmark & 6.82 & N/A & 0.3 \\
		& XNAS \cite{XNAS}  & NeurIPS19 & \checkmark & \checkmark & & \checkmark & 1.60 & 7.2 & 0.3\\
		& GDAS \cite{GDAS}  & CVPR19 & \checkmark & \checkmark & & \checkmark  & 2.82 &2.5 & 0.17 \\ 
		& FBNet-C \cite{Fbnet}  & CVPR19 & \checkmark & \checkmark& &\checkmark  & N/A & N/A & N/A \\ 
		&SGAS \cite{SGAS}  &CVPR20& \checkmark&\checkmark& &\checkmark  & 2.66 &3.7 & 0.25\\ 
		&GDAS-NSAS \cite{Overcoming Multi-Model Forgetting in One-Shot NAS with Diversity Maximization}&CVPR20  & \checkmark &\checkmark& &\checkmark  &2.73  & 3.5 & 0.4 \\ 
		& PC-DARTS $+$ Cutout \cite{PC-DARTS} &CVPR20& \checkmark&\checkmark& &\checkmark  &  2.57& 3.6 & 0.1 \\ 
		& R-DARTS \cite{R-DARTS} & ICLR20 & \checkmark&\checkmark& & \checkmark  &  2.93 & 4.1 & N/A \\ 
		\hline\hline 
		
		%RS
		\multirow{6}{*}{RS}& Hierarchical-EAS Random \cite{Hierarchical-EAS}&ICLR18  & \checkmark && &  &3.91 &N/A& 300 \\ 
		&NAO Random-WS \cite{NAO} &NeurIPS18 &\checkmark && & \checkmark & 3.92& 3.9& 0.3 \\ 
		&NASH-Net Random \cite{Simple and efficient architecture search for convolutional neural networks} &ICLR18& && &  & 6.50 & 4.4 & 0.2 \\ 
		& DARTS Random \cite{DARTS}&ICLR19&\checkmark && &  &3.29  &3.2  & 4 \\ 
		&\makecell[l]{RandomNAS $+$ Cutout \cite{RandomNAS}} &UAI19 & \checkmark&& &  \checkmark& 2.85& 4.3& 2.7  \\ 
		&RandomNAS-NSAS \cite{Overcoming Multi-Model Forgetting in One-Shot NAS with Diversity Maximization}&CVPR20  & \checkmark && &  &2.64  & 3.08 & 0.7 \\ 
		\hline\hline 
		
		%SMBO
		\multirow{6}{*}{SMBO} & NASBOT \cite{NASBOT}  &NeurIPS18& && &  & 8.69& N/A& 1.7  \\  
		&NAO \cite{NAO}&NeurIPS18  & \checkmark&\checkmark& &\checkmark  & 2.98 &28.6  &200  \\  
		&NAO-WS \cite{NAO}&NeurIPS18  & \checkmark&\checkmark& & \checkmark & 3.53 &2.5  &0.3  \\
		&NAO-Cutout \cite{NAO}  &NeurIPS18& \checkmark&\checkmark& & \checkmark & 2.11 &128  &200 \\
		&PNAS \cite{PNAS} &ECCV18 &\checkmark && & \checkmark &3.41  & 3.2 &225  \\ 
		&DPP-Net \cite{Dpp-net} &ECCV18 &\checkmark && & \checkmark &5.84  & 0.45 & 2 \\ 
		\hline
	\end{tabular}}
\end{table}

\begin{table}[!tp]
	\centering
	\caption{The performance comparison between the state-of-the-art NAS and mainstream artificial networks on ImageNet. The search strategy adopted by the corresponding NAS is consistent with Table \ref{tab:CIFAR-10}. Cutout is an augmentation technology used in \cite{cutout}.}
	\label{tab:ImageNet}
	\scriptsize
	\setlength{\tabcolsep}{1.8mm}{
	\begin{tabular}{|c|l|c|c|c|c|c|}
		\hline
		Search mothed                                & Reference                                                                              &Venue &\makecell[c]{Top 1/Top 5 \\Acc(\%)}             &\makecell[c]{ Params\\ (Millions) }        & \makecell[c]{Image Size\\ (squarred)}    & GPU Days                   \\ \hline\hline
		\multicolumn{1}{|c|}{\multirow{5}{*}{Human}} & \multicolumn{1}{l|}{Mobilenets \cite{Mobilenets} }                 &        CoRR17                                & \multicolumn{1}{c|}{70.6/89.5} & \multicolumn{1}{c|}{4.2}  & \multicolumn{1}{c|}{224} & \multicolumn{1}{c|}{-}     \\ 
		
		\multicolumn{1}{|c|}{}                       & \multicolumn{1}{l|}{ResNeXt \cite{ResNeXt} }        &CVPR17          & \multicolumn{1}{c|}{80.9/95.6} & \multicolumn{1}{c|}{83.6} & \multicolumn{1}{c|}{320} & \multicolumn{1}{c|}{-}     \\ 
		\multicolumn{1}{|c|}{}                       & \multicolumn{1}{l|}{Polynet \cite{Polynet} }&CVPR17                                              & \multicolumn{1}{c|}{81.3/95.8} & \multicolumn{1}{c|}{92.0} & \multicolumn{1}{c|}{331} & \multicolumn{1}{c|}{-}     \\ 
		\multicolumn{1}{|c|}{}                       & \multicolumn{1}{l|}{DPN \cite{DPN} }                                                &NIPS17 & \multicolumn{1}{c|}{81.5/95.8} & \multicolumn{1}{c|}{79.5} & \multicolumn{1}{c|}{320} & \multicolumn{1}{c|}{-}   \\
		
		\multicolumn{1}{|c|}{}                       & \multicolumn{1}{l|}{Shufflenet \cite{Shufflenet} }                  &CVPR18                                       & \multicolumn{1}{c|}{70.9/89.8} & \multicolumn{1}{c|}{5.0}  & \multicolumn{1}{c|}{224} & \multicolumn{1}{c|}{-}     \\
		\hline  \hline
		
		%RL
		\multicolumn{1}{|c|}{\multirow{6}{*}{RL}}                     & \multicolumn{1}{l|}{NASNet \cite{NASNet} }&CVPR18 & \multicolumn{1}{c|}{82.7/96.2} & \multicolumn{1}{c|}{88.9} & \multicolumn{1}{c|}{331} & \multicolumn{1}{c|}{2,000} \\ 
		& \multicolumn{1}{l|}{NASNet-A \cite{NASNet} }&CVPR18 & \multicolumn{1}{c|}{74.0/91.6} & \multicolumn{1}{c|}{5.3} & \multicolumn{1}{c|}{224} & \multicolumn{1}{c|}{2,000} \\ 
		\multicolumn{1}{|c|}{}                       & \multicolumn{1}{l|}{Block-QNN \cite{Block-QNN} }        &CVPR18           & \multicolumn{1}{c|}{77.4/93.5} & \multicolumn{1}{c|}{N/A}  & \multicolumn{1}{c|}{224} & \multicolumn{1}{c|}{96}    \\
		&N2N learning \cite{N2N learning}& ICLR18 & 69.8/N/A & 3.34 & 32 & 11.3 \\
		\multicolumn{1}{|c|}{}                       & \multicolumn{1}{l|}{Path-level EAS \cite{Path-level EAS} }       &ICML18                                           & \multicolumn{1}{c|}{74.6/91.9} & \multicolumn{1}{c|}{594}  & \multicolumn{1}{c|}{224} & \multicolumn{1}{c|}{200}   \\
		\multicolumn{1}{|c|}{}                       & \multicolumn{1}{l|}{FPNAS \cite{FPNAS} }       &ICCV19                                           & \multicolumn{1}{c|}{73.3/N/A} & \multicolumn{1}{c|}{3.41}  & \multicolumn{1}{c|}{224} & \multicolumn{1}{c|}{0.8}   \\
		\hline\hline
		
		%EA
		\multicolumn{1}{|c|}{\multirow{5}{*}{EA}}    & \multicolumn{1}{l|}{GeNet \cite{GeNet}}                        &ICCV17                                 & \multicolumn{1}{c|}{72.1/90.4} & \multicolumn{1}{c|}{156}  & \multicolumn{1}{c|}{224} & \multicolumn{1}{c|}{17}    \\ 
		\multicolumn{1}{|c|}{}                       & \multicolumn{1}{l|}{Hierarchical-EAS \cite{Hierarchical-EAS} } &ICLR18     & \multicolumn{1}{c|}{79.7/94.8} & \multicolumn{1}{c|}{64.0} & \multicolumn{1}{c|}{299} & \multicolumn{1}{c|}{300}   \\ 
		\multicolumn{1}{|c|}{}                       & \multicolumn{1}{l|}{AmoebaNet-A \textit{(N=6, F=190)} \cite{AmoebaNet-A} }         &AAAI19                                & \multicolumn{1}{c|}{82.8/96.1} & \multicolumn{1}{c|}{86.7} & \multicolumn{1}{c|}{331} & \multicolumn{1}{c|}{3,150} \\ 
		\multicolumn{1}{|c|}{}                       & \multicolumn{1}{l|}{AmoebaNet-A \textit{(N=6, F=448)} \cite{AmoebaNet-A}}                    &AAAI19                      & \multicolumn{1}{c|}{83.9/96.6} & \multicolumn{1}{c|}{469}  & \multicolumn{1}{c|}{331} & \multicolumn{1}{c|}{3,150} \\
		\multicolumn{1}{|c|}{}                       & \multicolumn{1}{l|}{Single-Path One-Shot NAS \cite{Single path one-shot neural architecture search with uniform sampling}}                   & CoRR19                      & \multicolumn{1}{c|}{74.7/N/A} & \multicolumn{1}{c|}{N/A}  & \multicolumn{1}{c|}{224} & \multicolumn{1}{c|}{\textless 1}
		\\ \hline\hline

%GO		
		\multicolumn{1}{|c|}{\multirow{16}{*}{GO}}    & \multicolumn{1}{l|}{Understanding One-Shot Models \cite{Understanding One-Shot Models} }    &ICML18       & \multicolumn{1}{c|}{75.2/N/A}  & \multicolumn{1}{c|}{11.9} & \multicolumn{1}{c|}{224} & \multicolumn{1}{c|}{N/A}   \\
		\multicolumn{1}{|c|}{}                       & \multicolumn{1}{l|}{SMASH \cite{SMASH} }            &ICLR18                                                  & \multicolumn{1}{c|}{61.4/83.7} & \multicolumn{1}{c|}{16.2} & \multicolumn{1}{c|}{32}  & \multicolumn{1}{c|}{3}     \\
		&\makecell[l]{Maskconnect \cite{Maskconnect}}& ECCV18  & 79.8/94.8 & N/A & 224 & N/A  \\ 
		\multicolumn{1}{|c|}{}                       & \multicolumn{1}{l|}{PARSEC \cite{PARSEC} }            &CoRR19                          & \multicolumn{1}{c|}{74.0/91.6} & \multicolumn{1}{c|}{5.6}  & \multicolumn{1}{c|}{N/A} & \multicolumn{1}{c|}{1}     \\ 
		\multicolumn{1}{|c|}{}                       & \multicolumn{1}{l|}{DARTS \cite{DARTS} }                    &ICLR19                                          & \multicolumn{1}{c|}{73.3/91.3} & \multicolumn{1}{c|}{4.7}  & \multicolumn{1}{c|}{224} & \multicolumn{1}{c|}{4}     \\ 
		\multicolumn{1}{|c|}{}                       & \multicolumn{1}{l|}{SNAS \cite{SNAS} }                                                              &ICLR19  & \multicolumn{1}{c|}{72.7/90.8} & \multicolumn{1}{c|}{4.3}  & \multicolumn{1}{c|}{224} & \multicolumn{1}{c|}{1.5}   \\
		\multicolumn{1}{|c|}{}                       & \multicolumn{1}{l|}{ProxylessNAS \cite{ProxylessNAS} }               &ICLR19                                         & \multicolumn{1}{c|}{75.1/92.5} & \multicolumn{1}{c|}{N/A}  & \multicolumn{1}{c|}{224} & \multicolumn{1}{c|}{8.33}  \\
		\multicolumn{1}{|c|}{}                       & \multicolumn{1}{l|}{GHN \cite{GHN} }                               &ICLR19  & \multicolumn{1}{c|}{73.0/91.3} & \multicolumn{1}{c|}{6.1}  & \multicolumn{1}{c|}{224} & \multicolumn{1}{c|}{0.84}  \\ 
		\multicolumn{1}{|c|}{}                       & \multicolumn{1}{l|}{SETN \cite{SETN} }                               & ICCV19  & \multicolumn{1}{c|}{74.3/92.0} & \multicolumn{1}{c|}{N/A}  & \multicolumn{1}{c|}{224} & \multicolumn{1}{c|}{1.8}  \\ 
		\multicolumn{1}{|c|}{}                       & \multicolumn{1}{l|}{TAS \cite{TAS} }                               & NeurIPS19  & \multicolumn{1}{c|}{69.2/89.2} & \multicolumn{1}{c|}{N/A}  & \multicolumn{1}{c|}{224} & \multicolumn{1}{c|}{2.5}  \\ 
		& XNAS \cite{XNAS}  & NeurIPS19 & 76.1/N/A & 5.2 & 224 & 0.3 \\
		\multicolumn{1}{|c|}{}                       & \multicolumn{1}{l|}{GDAS \cite{GDAS} }                               & CVPR19  & \multicolumn{1}{c|}{72.5/90.9} & \multicolumn{1}{c|}{4.4}  & \multicolumn{1}{c|}{224} & \multicolumn{1}{c|}{0.17}  \\ 
		& FBNet-C \cite{Fbnet}  & CVPR19 & 74.9/N/A & 5.5 & 224 & 9\\ 		
		\multicolumn{1}{|c|}{}                       & \multicolumn{1}{l|}{SGAS \cite{SGAS} }                               &CVPR20  & \multicolumn{1}{c|}{75.6/92.6} & \multicolumn{1}{c|}{5.4}  & \multicolumn{1}{c|}{224} & \multicolumn{1}{c|}{0.25}  \\ 
		\multicolumn{1}{|c|}{}                       & \multicolumn{1}{l|}{PC-DARTS \textit{(CIFAR10)} \cite{PC-DARTS} }                               &ICLR20  & \multicolumn{1}{c|}{74.9/92.2} & \multicolumn{1}{c|}{5.3}  & \multicolumn{1}{c|}{224} & \multicolumn{1}{c|}{0.1}  \\ \multicolumn{1}{|c|}{}                       & \multicolumn{1}{l|}{PC-DARTS \textit{(ImageNet)} \cite{PC-DARTS} }                               &ICLR20  & \multicolumn{1}{c|}{75.8/92.7} & \multicolumn{1}{c|}{5.3}  & \multicolumn{1}{c|}{224} & \multicolumn{1}{c|}{3.8}  \\ 
		\hline \hline
		
		%RS
		\multirow{1}{*}{RS}& Hierarchical-EAS Random \cite{Hierarchical-EAS}&ICLR18  &79.0/94.8 &N/A&299& 300 \\ 
		\hline \hline
		\multicolumn{1}{|c|}{\multirow{3}{*}{SMBO}}      & \multicolumn{1}{l|}{PNAS \textit{(Mobile)} \cite{PNAS} }           &ECCV18                                                    & \multicolumn{1}{c|}{74.2/91.9} & \multicolumn{1}{c|}{5.1}  & \multicolumn{1}{c|}{224} & \multicolumn{1}{c|}{225}   \\ 
		\multicolumn{1}{|c|}{}                       & \multicolumn{1}{l|}{PNAS \textit{(Large)} \cite{PNAS} }                                                               &ECCV18 & \multicolumn{1}{c|}{82.9/96.2} & \multicolumn{1}{c|}{86.1} & \multicolumn{1}{c|}{331} & \multicolumn{1}{c|}{225}   \\ 
		&DPP-Net \cite{Dpp-net} & ECCV18 & 75.8/92.9 & 77.2 & 224 & 2 \\
		\hline
	\end{tabular}}
\end{table}

The performance of the state-of-the-art NAS and mainstream artificial networks on the CIFAR-10 dataset are summarized in Table \ref{tab:CIFAR-10}. Moreover, based on Section \ref{sec:Optimization Strategy}, we simultaneously report the optimization strategy used in NAS. Similarly, we present a performance comparison on the ImageNet dataset in Table \ref{tab:ImageNet}. Because the optimization strategies used in the same NAS method are identical, we, therefore, omit the reporting of the corresponding optimization strategies in Table \ref{tab:ImageNet}.

From observing Table \ref{tab:CIFAR-10} and \ref{tab:ImageNet}, we can clearly see that in popular NAS, the use of modular search strategies is highly extensive. This is mainly because the modular search greatly reduces the complexity of the search space compared to a global search. However, as summarized in Section \ref{sec:Modular search space}, this does not necessarily imply that the modular search space is better than the global search. Moreover, we can conclude with certainty that incomplete training is also widely used; this can effectively accelerate the ranking of candidate neural architectures, thereby reducing the search duration. In addition, among the many optimization strategies considered, the gradient optimization based on the continuous search strategy can most significantly reduce the search cost and has a very rich body of research to support it. We also observe that random search strategy-based NAS has also achieved highly competitive performance. However, it is clear that existing NAS research on random search strategies is relatively inadequate. The optimization strategy of architecture recycling also exhibits comparatively excellent performance, but there are relatively few related studies. On the other hand,  transfer learning is a widely used technique in NAS tasks; that is, the architecture searched on a small dataset (CIFAR-10) will be transferred to a large dataset (ImageNet). Therefore, the search time costs involved in the two datasets shown in Table  \ref{tab:CIFAR-10} and \ref{tab:ImageNet} are the same \cite{DARTS,NASNet,FPNAS,SNAS}. These search tasks conducted in advance on small datasets are often called agent tasks. ProxylessNAS \cite{ProxylessNAS} also considers the problem of searching on large target tasks directly and without agents. 

In this part, we compare and summarize the popular NAS method with mainstream, artificially designed networks in multiple dimensions in terms of both performance and parameter amount. It is however worth noting that, the performance gains obtained using NAS are limited compared to those achieved using manually designed networks.

\section{Future Directions}
\label{sec:Future Directions}
The emergence of NAS is exciting, as it is expected to end the tedious trial and error process of manual neural architecture design. Moreover, it is also hoped that it can design a network structure that is completely different from the artificial network, thereby breaking through the existing human mindset.  Artificially designed networks have made breakthroughs in many fields, including image recognition \cite{VGG,Deep Subdomain Adaptation Network for Image Classification,Imagenet classification with deep convolutional neural networks}, machine translation \cite{LSTM,Neural Machine Translation With GRU-Gated Attention Model,Google's neural machine translation system,BERT}, semantic segmentation \cite{Fully convolutional networks for semantic segmentation,Cocoon Image Segmentation Method Based on Fully Convolutional Networks,Evolutionary Compression of Deep Neural Networks for Biomedical Image Segmentation}, object detection \cite{Faster r-cnn,From Discriminant to Complete,Real-Time Object Detection With Reduced Region Proposal Network via Multi-Feature Concatenation}, and video understanding \cite{MASK-RL,Deep Attentive Video Summarization With Distribution Consistency Learning} and so on.  Although the NAS-related research has been quite rich, compared with the rapid development of artificial neural architecture design in various fields, NAS is still in the preliminary research stage.  The current NAS is primarily focused on improving image classification accuracy, compressing the time required to search for neural architecture, and making it as light and easy to promote as possible.  In this process, a proper baseline is crucial, as this helps to avoid NAS search strategy research being masked by various powerful augmenting technologies, regularization, and search space design tricks. Besides, the search strategies currently used by NAS are relatively concentrated, especially GO based on supernets, and there are many theoretical deficiencies in the related research. Therefore, improving the related research background is of great benefit to the development of NAS.

More specifically, from the original intention of NAS design, early NAS was very close to people's expectations for automatic neural architecture design.  For example, Large-scale Evolution \cite{Large-scale Evolution} uses an EA as a NAS search strategy, and emphasizes that there is no need for manual participation once network evolution begins. In the case of reducing human interference as much as possible, the algorithm is allowed to autonomously determine the evolution direction of the network. This is a good start, although the search performance at the time was insufficient. This is mainly because Large-scale Evolution emphasizes the reduction of artificial restrictions to the greatest extent possible, and also evolves from the simplest single-layer network (as shown in Fig.\ref{fig:Search_from_scratch}). This means that the population must contain a sufficient number of individuals to evolve satisfactory results. 
%This means that the evolution process consumes a lot of computing resources, which increases the possibility overall of evolving a network structure that is free from human inherent thinking. 
Besides, due to the limitations of evolutionary efficiency and the huge search space, it is difficult to evolve a network structure that can achieve outstanding performance. Therefore, subsequent NAS works began to discuss how to reduce the search space as much as possible while also improving network performance. NASNet \cite{NASNet} benefited from the idea of artificial neural architecture design \cite{VGG,Resnet,GoogLeNet}, and proposes a modular search space that was later widely adopted. However, this comes at the expense of the freedom of neural architecture design. Although this modular search space does significantly reduce the search cost, it is unclear in reality whether a better neural architecture has been ignored in the process of turning global search into modular search. This is also the main reason why it is unclear whether the modular search is better than a global search (related research is also lacking).  Besides, the freedom of neural architecture design and the search cost presents a contradiction. How to balance the two and obtain good performance remains an important future research direction. 

Also, there are two issues worthy of vigilance. From the perspective of the baseline, as analyzed in Section \ref{sec:Performance comparison}, a widely criticized problem of NAS is the lack of a corresponding baseline and sharable experimental protocol, which makes it difficult to compare NAS search algorithms with each other.  While RS has proven to be a strong baseline \cite{Evaluating the search phase of neural architecture search,NAS evaluation is frustratingly hard,Hierarchical-EAS,RandomNAS}, there is still insufficient research on this subject. \cite{Evaluating the search phase of neural architecture search} pointed out that the performance of the current optimal NAS algorithm is similar to the random strategy, which should arouse researchers' concerns.  Therefore, more ablation experiments are necessary, and researchers should pay more attention to which part of the NAS design leads to performance gains.  Blindly stacking certain techniques to increase performance should also be criticized. Therefore, relevant theoretical analysis and reflection are crucial to the future development of NAS. 

Another issue that requires vigilant is the widely used parameter sharing strategy \cite{ENAS}; although it effectively improves NAS search efficiency, an increasing body of evidence shows that the parameter sharing strategy is likely to result in a sub-optimal inaccurate candidate architecture ranking \cite{Blockwisely Supervised Neural Architecture Search with Knowledge Distillation,Evaluating the search phase of neural architecture search,Overcoming Multi-Model Forgetting in One-Shot NAS with Diversity Maximization,Once for all}. This would make it almost impossible for NAS to find the optimal neural architecture in the search space.  Therefore, relevant theoretical research and improvements are expected in the future. Besides, the current NAS works are mainly focused on improving the accuracy of image classification and reducing the search cost. In the future, NAS will play a greater role in areas that require complex neural architecture design, such as multi-object architecture search, network transformation that employs NAS based on existing models, model compression, target detection, and segmentation, and so on.

RobNet \cite{RobNet} uses the NAS method to generate a large number of neural architectures by analyzing the differences between strongly and poorly performing architectures; this will enable the identification of which network structures help to obtain a robust neural architecture. Inspired by this, the once feasible idea would be to use a similar concept to RobNet in a search space with a higher degree of freedom, analyze the structural features common to promising architectures, and increase their structural proportion in the search space. Besides, it enables a reduction in the proportion of the common structural features of poorly performing architectures in the search space, to gradually reduce the search space in stages. The algorithm should be allowed to autonomously decide which network structures should be removed or added rather than artificially imposing constraints.

On the other hand, in the design of deep neural networks, in addition to the design of the neural architecture that consumes a lot of researchers' energy, another headache is the setting of non-architecture hyperparameters (for example optimizer type, initial learning rate, weight decay, mixup ratio \cite{mixup}, drop out ratio and stochastic depth drop ratio \cite{stochastic}, etc.). Because the impact of these hyperparameters on network performance is also crucial. Moreover, the number of possible related non-architectural hyperparameter groups and architectural combinations is often very large, and it is difficult and inefficient to manually design all possible combinations. Therefore, it is also a very promising research direction to consider how to integrate the hyperparameters in the neural architecture into the search space and allow the algorithm to automatically adjust the settings of the hyperparameters. Although some early methods also explored the joint search of hyperparameters and architectures, they mainly focused on small data sets and small search spaces. Fortunately, the experiments of FBNetV3 \cite{FBNetV3} and AutoHAS \cite{AutoHAS} show that this direction has great potential, and it is expected to find higher-accuracy architecture-recipe combinations.

In short, the emergence of NAS is an exciting development. At present, NAS is still in its initial stages, and additional theoretical guidance and experimental analysis are required. Determining which NAS designs lead to improved performance is a critical element of improving NAS.  If NAS is to completely replace the design of artificial neural architecture, more research, and a more solid theoretical basis are required. There is still a long way to go.

\section{Review Threats}
\label{sec:Review Threats}
We made the above summary and discussion based on the open NAS related work. We objectively state the viewpoints in some unpublished papers that need to be further confirmed. Due to the limited time and energy, we directly applied the results in the references without reproducing their results.

\section{Summary and Conclusions}
\label{sec:Summary and Conclusions}
This survey provides an insightful observation of the early NAS work and summarizes the four characteristics shared by the early NAS. In response to these challenges, we conducted a comprehensive and systematic analysis and comparison of existing work in turn. For the first time, we reviewed all NAS-related work from the perspective of challenges and corresponding solutions. And the performance of existing NAS work was comprehensively compared and discussed. Finally, we conducted a comprehensive discussion on the possible future direction of NAS and raised two questions worthy of vigilance. This is very useful for future NAS research.

%%
%% The acknowledgments section is defined using the "acks" environment
%% (and NOT an unnumbered section). This ensures the proper
%% identification of the section in the article metadata, and the
%% consistent spelling of the heading.
\begin{acks}
This work was partially supported by the NSFC under Grant (No.61972315 and No. 62072372), the Shaanxi Science and Technology Innovation Team Support Project under grant agreement (No.2018TD-026) and the Australian Research Council Discovery Early Career Researcher Award (No.DE190100626).
\end{acks}

%%
%% The next two lines define the bibliography style to be used, and
%% the bibliography file.
%\bibliographystyle{ACM-Reference-Format}
%\bibliography{sample-base}

\end{document}